\documentclass[10pt,twocolumn,letterpaper]{article}
\usepackage{amsmath}
\usepackage{amssymb}
\usepackage{graphicx}
\usepackage{algorithm}
\usepackage{algpseudocode}
\usepackage{xcolor}
\usepackage{wacv}      

\IfFileExists{axessibility.sty}{\usepackage[accsupp]{axessibility}}{}

\definecolor{wacvblue}{rgb}{0.21,0.49,0.74}
\usepackage[pagebackref,breaklinks,colorlinks,allcolors=wacvblue]{hyperref}

\def\wacvPaperID{556} 
\def\confName{WACV}
\def\confYear{2027}

\title{Grounding Image Composition with Label-Conditioned Visual Evidence}

\author{%
{\normalsize Armin Dadras\textsuperscript{1,2}\quad Tingyu Lin\textsuperscript{1}\quad Robert Sablatnig\textsuperscript{1}}\\
{\normalsize Franziska Proksa\textsuperscript{2}\quad Markus Seidl\textsuperscript{2}}\\[2pt]
{\small \textsuperscript{1}Computer Vision Lab, TU Wien, Vienna, Austria}\\
{\small \textsuperscript{2}UAS St. P\"olten, St. P\"olten, Austria}\\
{\tt\small adadras@cvl.tuwien.ac.at}
}

\begin{document}
\maketitle
\begin{abstract}
Image composition labels such as \emph{center}, \emph{diagonal}, or \emph{rule
of thirds} describe visual structure rather than object categories. KU-PCP
provides image-level labels and class-specific geometric elements, but does
not show which visible regions people associate with each of its nine classes. We introduce KU-PCP
Elements (KU-PCPE), a label-conditioned region-evidence benchmark for
photographic composition. Given an image and a label already judged
applicable, the task is to segment the visible regions that explain why the
label applies. KU-PCPE contains 4,093 single-annotator training masks and a
1,327-task expert test split with three masks per image-label task. The annotations
also include exploratory segment-versus-line ratings that describe
whether region or semantic-guideline evidence better explains the label. A
held-out expert reaches .867 Dice against the other experts, while the
strongest deployable model reaches .664, a proposal oracle .672, and the best
vision-language model (VLM) proposal selector .395. KU-PCPE exposes the gap
between recognizing a
composition label and grounding the human-aligned visual evidence that makes
it plausible in a particular image.
\end{abstract}

\section{Introduction}
\label{sec:introduction}

Photographic composition is an inherently visual concept. Labels such as
center, diagonal, horizontal, or rule of thirds name recurring spatial
organizations, but they do not specify which visible structures make the label
applicable in a particular image. Two images may share a label for different
reasons, and one image may support several labels through different regions,
lines, or partitions. This creates an evidence question that image-level
classification cannot answer: given an image and a queried composition label,
what visual evidence makes that label plausible?

We introduce KU-PCP Elements (KU-PCPE), a label-conditioned region-evidence
grounding benchmark for photographic composition. Each task pairs a photograph
with one of its KU-PCP composition labels, and annotators select the visible
regions they consider jointly responsible for that composition. A mask may
contain one or several disconnected regions and may include objects, object
parts, background, or negative space. For labels expressed through a line,
axis, or spatial relation, the mask contains the visible elements or partitions
that carry the relation. A multi-label photograph therefore yields a distinct
grounding task for each of its composition labels.

KU-PCPE builds on KU-PCP but provides a different target. KU-PCP supplies
image-level composition labels and class-specific geometric elements such as
subject boxes, leading lines, symmetry axes, triangles, and sky regions
\cite{lee2018kupcp}. KU-PCPE provides one human region representation across
all nine classes and three expert interpretations for each test task. A limited
comparison with KU-PCP's detected triangles shows shared geometry, while our
masks also capture the objects, surfaces, and background that make the
composition visible.

\begin{figure*}[t]
\centering
\includegraphics[width=0.98\textwidth]{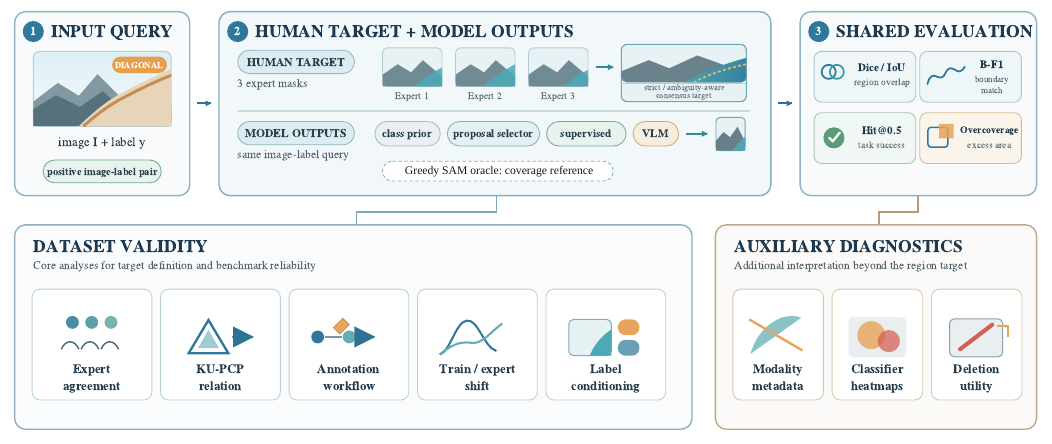}
\caption{Overview of KU-PCPE. The primary benchmark predicts which visible regions explain an applicable
composition label in a given image.
Different method families share the target and evaluation. The remaining
results either establish the validity of the dataset and target or provide
supporting diagnostics of what the masks capture.}
\label{fig:benchmark_overview}
\end{figure*}

KU-PCPE has one primary benchmark: predicting region evidence for an image and
an applicable composition label. Class priors, proposal selectors, supervised
segmenters, and vision-language model (VLM) selectors are alternative ways to
produce the same output, scored against the same expert consensus. Proposal
oracles measure how well candidate masks can cover that target. The agreement
studies and dataset analyses establish what the target represents and how
reliably it can be measured. Modality ratings characterize explanatory form,
while classifier heatmaps and deletion experiments connect the masks to
image-level recognition.
\Cref{fig:benchmark_overview} summarizes this organization.

Our contributions are:
\begin{itemize}
\item KU-PCPE, a label-conditioned region-evidence dataset with
4,093 training masks and 1,327 three-expert test tasks across nine composition
classes;
\item a common region target, expert agreement analysis, and studies of label
conditioning, annotation workflow, annotator variation, and the relationship
to KU-PCP's geometric elements; and
\item a region-evidence benchmark with model-free, proposal-based, supervised,
and VLM baselines, complemented by recognition diagnostics.
\end{itemize}
The dataset, evaluation code, and reconstruction tools are available at
\href{https://adadras.github.io/ku-pcpe/}{\nolinkurl{adadras.github.io/ku-pcpe/}}.

\section{Related Work}
\label{sec:related}

\paragraph{Composition labels and assessment.}
Most computational work treats photographic composition as a global label,
score, or embedding. Broad aesthetics datasets such as AVA and AADB provide
aesthetic ratings and attributes, while composition-specific
resources such as KU-PCP, CADB, LODB, and PICD evaluate composition classes,
quality, retrieval, or learned representations
\cite{murray2012ava,kong2016aadb,lee2018kupcp,zhang2021cadb,zhao2024lodb,zhao2025picd}.
These resources establish useful composition vocabularies and assessment
protocols, but they do not provide a common human region target for a queried
label. KU-PCPE adds that target at the image-label level.

\paragraph{Geometric composition elements.}
Several approaches make composition visible through geometric or pattern-based
structure. KU-PCP uses class-specific automatic elements, including
main-subject boxes, leading lines, a symmetry axis, triangles, and sky regions
\cite{lee2018kupcp}. Triangle composition has also been modeled with
handcrafted line and geometric descriptors for retrieval and feedback
\cite{zhou2016triangles}, while SAMP-Net learns scores over predefined
composition partitions~\cite{zhang2021cadb}. These representations make
selected composition cues explicit. KU-PCPE instead asks people to select
visible region support using one representation across all nine labels. We
compare the overlapping triangle representation directly in
\Cref{sec:kupcp_relation}.

\paragraph{Semantic lines and layout primitives.}
Semantic-line detection makes one important composition primitive explicit by
detecting visually meaningful lines that separate semantic regions or convey
layout structure \cite{lee2017semanticline,han2020dht,jin2020semanticline,ko2024slcd}.
Recent detectors improve line selection by modeling candidate ranking,
pairwise relations, or globally harmonious line combinations, and have been
used for horizon, vanishing-point, symmetry, and composition-based retrieval
tasks. Yet line-based representations remain strongest for straight
guidelines such as horizons, diagonals, perspective, and symmetry axes. They
are less suitable for curved layouts, repeated patterns, negative space, or
region-ordering effects. KU-PCPE therefore treats semantic lines as one possible
explanatory modality rather than as complete composition evidence.

\paragraph{Visual and textual explanations.}
Post-hoc visual explanation methods such as Grad-CAM localize classifier
evidence, but they explain a trained model's decision rather than human-marked
composition evidence for a queried label \cite{selvaraju2017gradcam}. Recent
aesthetic and composition work has also moved toward language: neural
reviewers generate comments, photo-critique datasets pair images with textual
feedback, VILA learns aesthetic representations from user comments, and MICA
jointly predicts composition scores and explanations
\cite{wang2019neuralreviewer,nieto2022rpcd,ke2023vila,yuan2026mica}. These
works show that scores and labels are insufficient, but textual critiques can
say that an image is balanced, centered, or diagonal without localizing the
responsible region, partition, or axis. Since composition is inherently
spatial, KU-PCPE complements textual explanations with masks showing which
regions support a given label. It also records whether annotators prefer
regions or lines as explanations.

\section{KU-PCP Elements Dataset}
\label{sec:dataset}

\subsection{Evidence Target}

KU-PCPE presents each task as a KU-PCP image paired with one
of its composition labels. The target contains the visible regions an
annotator considers jointly responsible for that composition. Evidence may
contain one or several disconnected regions and may include objects, object
parts, background, or negative space. For compositions expressed through a
line, axis, or spatial relation, the mask contains the visible elements or
partitions that carry the relation. Binary masks thus serve all nine classes
while allowing experts to choose different regions.

The labels are horizontal, vertical, diagonal, curved, triangle, rule of
thirds, center, symmetric, and pattern. Multi-label images contribute one task
per KU-PCP label, so each label is grounded in its own visible evidence. The
supplementary material provides the full class guide and both representative
and ambiguous examples for every class.

\begin{table}[t]
\centering
\scriptsize
\setlength{\tabcolsep}{4pt}
\begin{tabular}{lrr}
\toprule
Class & Train tasks & Expert-test tasks \\
\midrule
Horizontal & 758 & 211 \\
Vertical & 195 & 66 \\
Diagonal & 471 & 189 \\
Curved & 235 & 92 \\
Triangle & 374 & 171 \\
Rule of thirds & 923 & 125 \\
Center & 750 & 311 \\
Symmetric & 232 & 99 \\
Pattern & 155 & 63 \\
\midrule
\textbf{Total} & \textbf{4,093} & \textbf{1,327} \\
\bottomrule
\end{tabular}
\caption{Image-label tasks. The single-annotator train split contains
3,169 images and provides training supervision. The expert-test split contains
1,073 images and three masks per task.}
\label{tab:class_distribution}
\end{table}

Class frequencies vary substantially in both splits
(\Cref{tab:class_distribution}). The training split provides development supervision. Human agreement and
benchmark evaluation are defined on the three-expert test split.

\subsection{Annotation Protocol}
\label{sec:annotation_process}

\begin{figure*}[t]
\centering
\begin{minipage}[t]{0.49\textwidth}
\centering
\includegraphics[width=\linewidth,height=0.24\textheight,keepaspectratio]{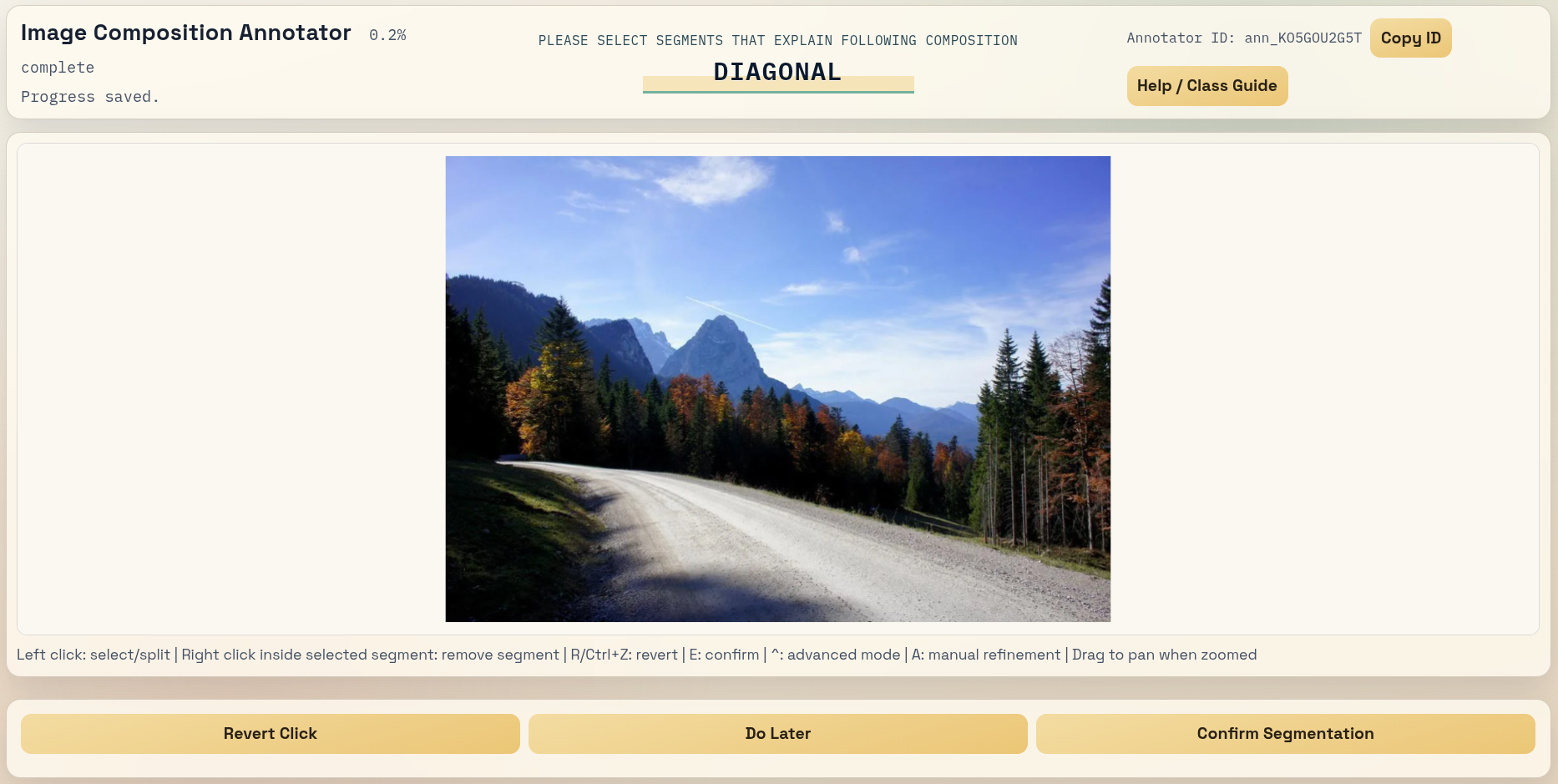}\\[-0.35em]
\small (a) Region selection and mask confirmation.
\end{minipage}\hfill
\begin{minipage}[t]{0.49\textwidth}
\centering
\includegraphics[width=\linewidth,height=0.24\textheight,keepaspectratio]{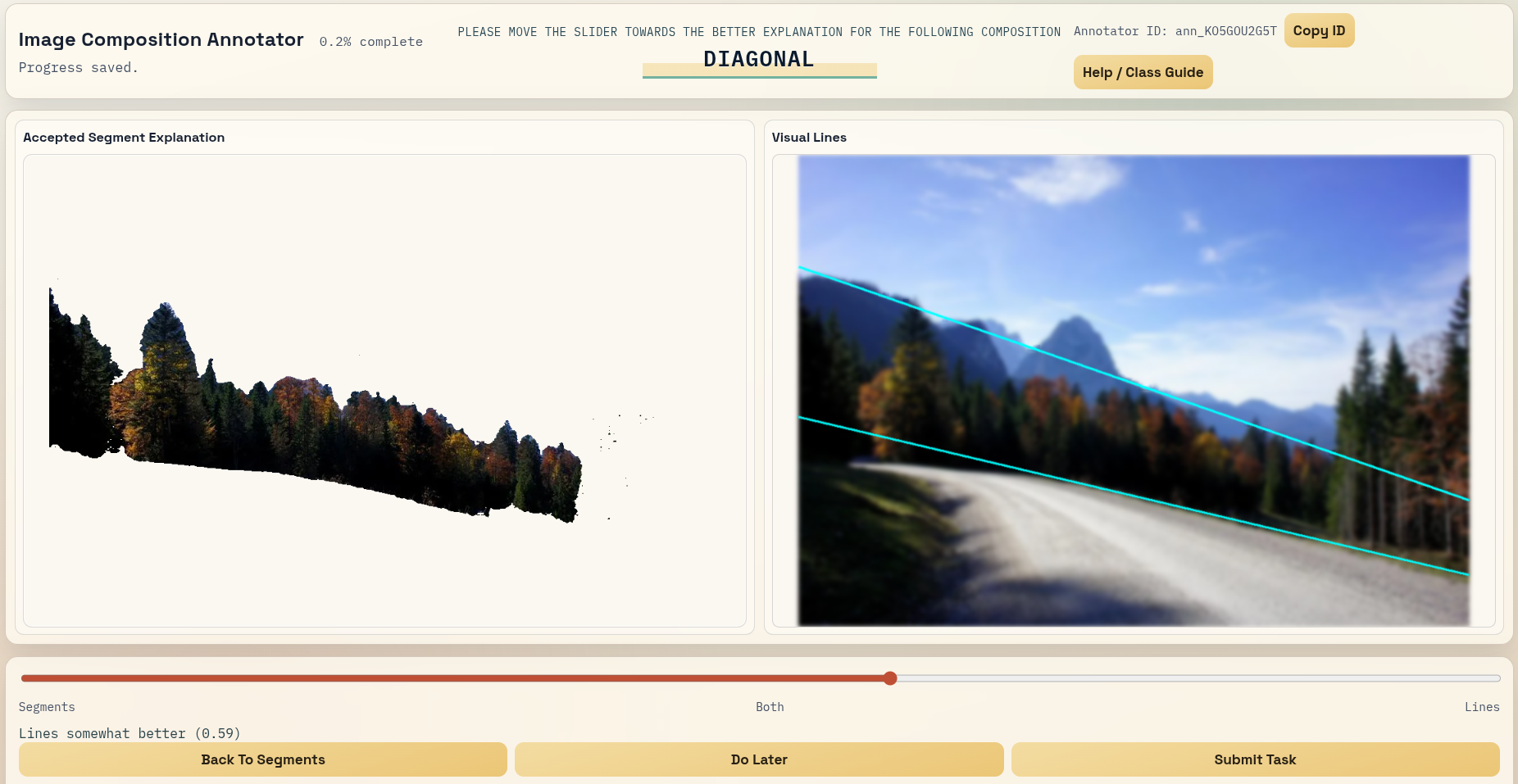}\\[-0.35em]
\small (b) Exploratory segment-versus-line rating.
\end{minipage}
\caption{KU-PCPE annotation interface. Annotators select or refine visible
region evidence for the queried composition and then compare the confirmed
mask with an automatic semantic-line overlay.}
\label{fig:annotation_interface}
\end{figure*}

Annotators first saw an image and its queried composition label. Precomputed
candidate regions were revealed by clicking in the image
(\Cref{fig:annotation_interface}a). They could select several regions, add or
split regions, remove selections, and manually refine the final mask. The
candidates were generated with SAM2 Hiera-L~\cite{ravi2024sam2}. The candidates reduced the effort of drawing boundaries, while the annotator decided which visible
regions supported the label and confirmed the result. Annotators worked
independently and did not see one another's masks. They received written instructions and a guide defining all nine classes,
with three visual cues per class (see the supplement).

The three test annotators all had formal education in arts or design and more
than five years of professional experience in art, design, photography, or
professional visual practice. A separate annotator produced the training split
with the same task and interface. Records from all four annotators show that
most annotations had no geometry action logged (\Cref{tab:annotation_workflow}).
\Cref{sec:workflow_results} combines the three experts' retained records at the task level.

\begin{table}[t]
\centering
\scriptsize
\setlength{\tabcolsep}{3pt}
\begin{tabular}{lrr}
\toprule
Interface statistic & Train & Expert test \\
\midrule
Annotations with records & 4,093 & 3,981 \\
No geometry action & 51.8\% & 62.2\% \\
Any geometry action & 48.2\% & 37.8\% \\
Manual tool use & 15.1\% & 17.5\% \\
Split operation & 5.7\% & 6.1\% \\
Median recorded actions & 15 & 8 \\
\bottomrule
\end{tabular}
\caption{Annotation workflow across all four annotators. Test percentages and medians pool 3,981 annotations from three experts on 1,327 tasks; training has one annotation per task. Manual tool use includes entering, applying, exiting, or undoing refinement. Per-annotator results are in the supplement.}
\label{tab:annotation_workflow}
\end{table}

The interface also displayed automatically detected semantic lines. After
confirming a mask, annotators could rate whether the region mask or the line
overlay better explained the queried label (\Cref{fig:annotation_interface}b). The overlay served as an
annotation aid, while the confirmed region mask defined the binary target.

\subsection{Expert Agreement}
\label{sec:expert_agreement}

We have three masks for every test image and applicable label. Their
pixelwise majority defines the \emph{strict} target. The
\emph{ambiguity-aware} target complements masks whose opposite side agrees
better with the first expert before taking the majority. This captures
partition ambiguities such as selecting sky or ground around the same horizon.
Both targets are included in the annotation package and reported.

\begin{figure*}[t]
\centering
\includegraphics[width=\textwidth]{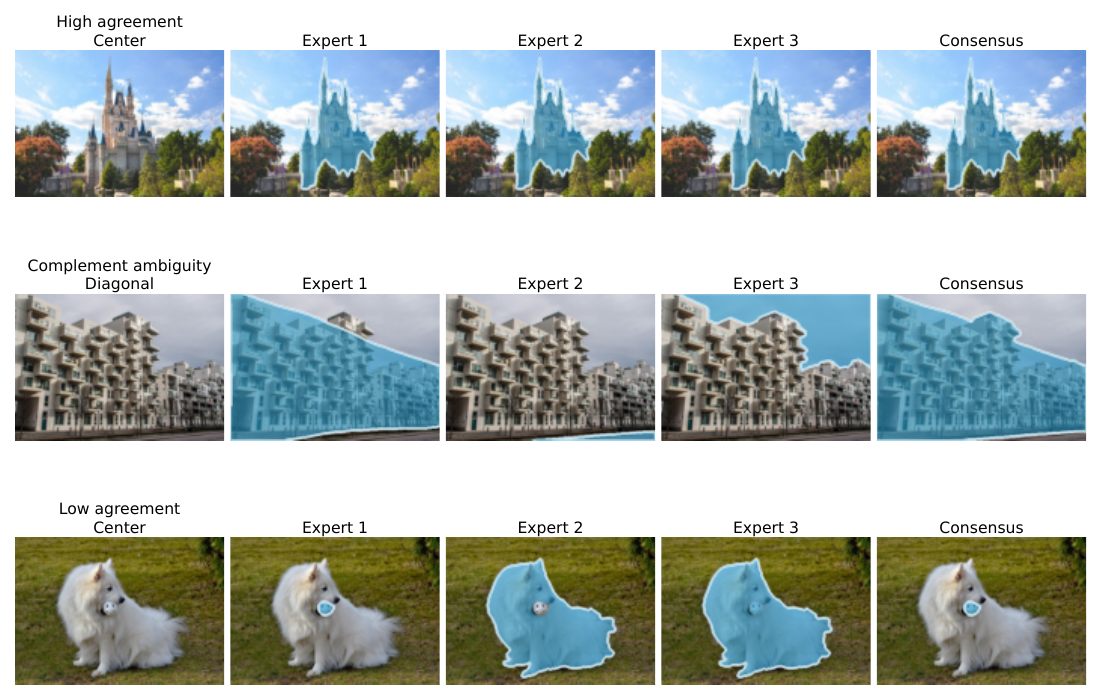}
\caption{Expert agreement examples. Rows show high agreement, complementary
region ambiguity, and low agreement. Columns show the image, three expert
masks, and the aligned majority. Disagreement often reflects alternative
plausible supports rather than small boundary differences.}
\label{fig:expert_agreement_examples}
\end{figure*}

Mean raw pairwise Dice is .7428. A held-out expert reaches Dice .8666,
intersection-over-union (IoU) .8054, boundary F1 .7541, and Hit@0.5 .9312
against the aligned consensus of the other two experts.
\Cref{tab:per_class_agreement} shows strongest agreement for rule of thirds,
center, and triangle. Horizontal, diagonal, curved, and pattern admit more
complementary or alternative readings, as illustrated in
\Cref{fig:expert_agreement_examples}.

\begin{table}[t]
\centering
\scriptsize
\setlength{\tabcolsep}{2.4pt}
\begin{tabular}{lrrrrr}
\toprule
Class & Tasks & Pair D & Maj. D & $\kappa$ & Flip \\
\midrule
Horizontal & 211 & .6371 & .9539 & .8281 & .5118 \\
Vertical & 66 & .6557 & .8362 & .5782 & .1515 \\
Diagonal & 189 & .5973 & .8805 & .5433 & .4656 \\
Curved & 92 & .6105 & .8556 & .5240 & .3696 \\
Triangle & 171 & .8332 & .9351 & .7796 & .1170 \\
Rule of thirds & 125 & .9538 & .9782 & .9426 & .0080 \\
Center & 311 & .8427 & .9235 & .8014 & .0354 \\
Symmetric & 99 & .8239 & .9165 & .7152 & .0505 \\
Pattern & 63 & .5324 & .8092 & .2510 & .5397 \\
\bottomrule
\end{tabular}
\caption{Expert agreement by class. Pair D is raw pairwise Dice; Maj. D is
Dice to the jointly aligned majority, with expert 1 fixed and the remaining
orientations maximizing mean pairwise Dice. Flip is the fraction
of tasks for which at least one complement alignment is used. Full metrics and
confidence intervals are in the supplement.}
\label{tab:per_class_agreement}
\end{table}

\subsection{Exploratory Evidence Modality}
\label{sec:evidence_modality}

\begin{figure}[t]
\centering
\includegraphics[width=\linewidth]{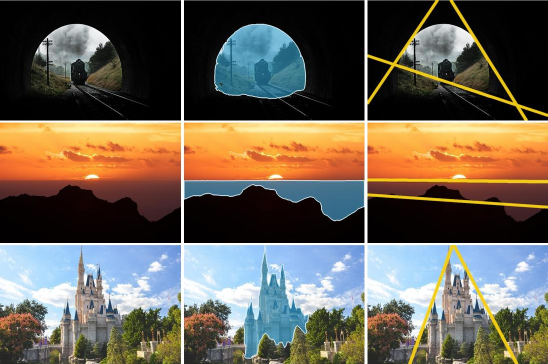}
\caption{Exploratory modality examples. Top: segment-preferred center
($r=.22$); middle: mixed horizontal ($r=.50$); bottom: line-preferred triangle
($r=.81$). Each row shows the image, human region evidence, and automatic
semantic-line overlay.}
\label{fig:evidence_modality_examples}
\end{figure}

\begin{figure}[t]
\centering
\includegraphics[width=\linewidth]{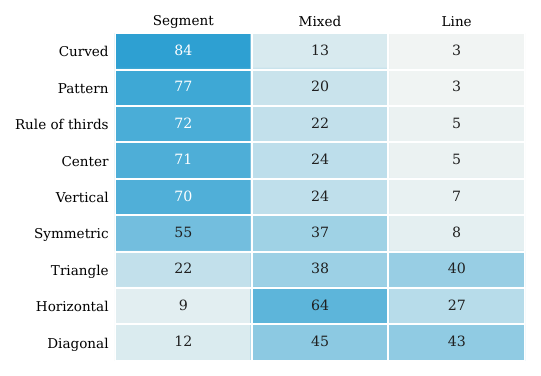}
\caption{Evidence modality by composition class. Cells give the percentage of
expert-test ratings that prefer segments, mixed evidence, or lines. Full
values are in the supplement.}
\label{fig:evidence_modality_by_class}
\end{figure}

Experts also rated whether the confirmed region mask or an automatic
semantic-line overlay better explained the queried label, using a score
$r\in[0,1]$ from segment to line preference. After binning ratings as segment,
mixed, or line, pairwise agreement is .54 and Fleiss $\kappa=.26$; continuous
ratings have mean pairwise $|\Delta r|=.22$. These exploratory ratings describe
whether experts prefer regions or lines as explanations.
The three preferences appear in \Cref{fig:evidence_modality_examples}.
Horizontal is mainly mixed, diagonal and triangle have the largest
line-preferred shares, and most other classes are segment-dominated
(\Cref{fig:evidence_modality_by_class}).

\section{Benchmark Tasks}
\label{sec:tasks}

Evidence segmentation is the primary benchmark task. The prediction protocols
described below differ in how they produce the same binary target; proposal
oracles measure candidate coverage, while modality metadata remains
exploratory.

\paragraph{Evidence segmentation.}
Given an image $I$ and a composition label $y$ assigned to that image in
KU-PCP, a method predicts a binary evidence mask $M$. Unless stated otherwise,
evaluation uses the ambiguity-aware expert consensus on the test split.

\paragraph{Proposal coverage and selection.}
SAM2.1 Hiera-L produces a candidate set for each image.
Greedy Oracle@$k$ measures how much of the target can be covered when the
reference mask is available. Random, area-based, and class-template selectors
choose without access to the reference and are deployable controls.

\paragraph{Supervised segmentation.}
Image-only and label-conditioned U-Net~\cite{ronneberger2015unet} variants and a label-conditioned
SegFormer-B0~\cite{xie2021segformer} learn from masks provided by one training annotator; label-conditioned models also receive the
queried composition label.

\paragraph{VLM proposal selection.}
For each task, a VLM sees the queried label and an overlay with up to 12
numbered proposals, then returns up to three IDs. Their union is evaluated as
a binary mask. This yields a language-conditioned proposal selector evaluated
against the expert region target.

\paragraph{Exploratory modality metadata.}
Segment-versus-line prediction uses the expert ratings as an auxiliary
diagnostic. Full VLM modality results are reported in the supplement.

\section{Experiments}
\label{sec:experiments}

We organize the results in three stages. First, we examine the dataset and
target through agreement, evaluation sensitivity, the relation to KU-PCP,
annotation workflow, annotator variation, and label conditioning. Next, all
predictive methods are compared on the same region-evidence task and metrics;
proposal oracles report candidate coverage. Finally, classifier explanations
and deletion connect the annotations to image-level recognition as auxiliary
analyses.

\subsection{Benchmark Protocol}

The proposal experiments use SAM2.1 Hiera-L. Supervised
baselines comprise an image-only U-Net, a label-conditioned U-Net, and a
label-conditioned SegFormer-B0. Model-free baselines include
random matched-area masks and class templates. We also evaluate a greedy
proposal oracle and a deployable template selector. The supplement specifies
the training, proposal, prompting, and thresholding settings. The recognition
diagnostics use a ResNet-50 composition classifier, Grad-CAM explanations, and
DINO ViT patch-similarity maps
\cite{he2016resnet,selvaraju2017gradcam,caron2021dino}.
The VLM suite covers Qwen3-VL, InternVL3.5, MiniCPM-V 4.5, Molmo2, and
LLaVA-OneVision~\cite{qwen2025qwen3vl,wang2025internvl35,minicpm2025v45,molmo2026molmo2,li2024llavaonevision};
exact checkpoints are listed in the supplement.

\paragraph{Metrics.}
Dice and IoU measure region overlap. Boundary F1 (B-F1) uses a two-pixel
tolerance after resizing the longer image side to at most 256 pixels.
Overcoverage is the fraction of a predicted mask that lies outside the target;
lower is better. Hit@0.5 is the fraction of tasks with Dice at least .5. Main
results use the ambiguity-aware target. Strict results and 95\% bootstrap
confidence intervals over image-label tasks are reported in the supplement.

\paragraph{VLM responses.}
The prompt shows the queried label and up to 12 numbered proposals and asks for
a JSON list of at most three IDs. We keep the first three distinct valid IDs;
duplicates and unavailable IDs are ignored, and a response with no valid ID
produces an empty mask. Decoding is deterministic. Exact prompts, parsing
rules, and invalid-response rates are provided in the supplement.

\subsection{Evaluation Sensitivity}
\label{sec:evaluation_sensitivity}

Sensitivity analysis uses both strict and ambiguity-aware targets. The comparison in
\Cref{tab:target_sensitivity} shows that the central ordering is
unchanged. Oracle@3 is selected separately for each target; all other
predictions are fixed. Full metrics and confidence intervals are reported in
the supplement.

\begin{table}[t]
\centering
\scriptsize
\setlength{\tabcolsep}{5pt}
\begin{tabular}{lrr}
\toprule
Method & Strict Dice & Ambig. Dice \\
\midrule
Train class template & .451 & .462 \\
SAM2.1 Oracle@3 & .679 & .672 \\
SegFormer-B0 & .662 & .664 \\
Qwen3-VL-32B & .399 & .395 \\
\bottomrule
\end{tabular}
\caption{Sensitivity to target construction. The benchmark pattern is stable
under strict and ambiguity-aware evaluation.}
\label{tab:target_sensitivity}
\end{table}

\subsection{Relation to Original KU-PCP}
\label{sec:kupcp_relation}

KU-PCP's local elements span boxes, lines, axes, triangles, and sky regions.
The 171 triangle tasks with matched KU-PCP detector output provide a direct
comparison between a geometric element and the KU-PCPE region target. We
compare both the detected sides and the image region they enclose with the
KU-PCPE targets.

\begin{table}[t]
\centering
\scriptsize
\setlength{\tabcolsep}{4pt}
\begin{tabular}{lrr}
\toprule
Triangle comparison & Strict & Ambig. \\
\midrule
Sides inside target & .374 & .398 \\
Sides inside target, 5-px tolerance & .465 & .489 \\
Enclosed region Dice & .572 & .585 \\
Enclosed region IoU & .451 & .462 \\
\bottomrule
\end{tabular}
\caption{Comparison with KU-PCP triangle detector output on 171 matched tasks.
The first two rows give the fraction of detected side pixels supported by the
expert target; the last two compare the enclosed region.}
\label{tab:kupcp_triangle}
\end{table}

The partial overlap in \Cref{tab:kupcp_triangle} confirms that the geometric triangle and the human region
target describe related structure. The human region target additionally
captures the objects, surfaces, and background regions through which the
triangle is perceived. Details and examples are in the supplement.

\subsection{Results by Annotation Workflow}
\label{sec:workflow_results}

We combine the three experts' retained interaction records. A test task is
in the no-edit group only when none of its experts recorded a geometry action;
it is in the edit group when at least one expert did. Agreement uses all three
final masks, and model scores use the ambiguity-aware target. Each of the
1,327 image-label tasks contributes once.

\begin{table}[t]
\centering
\scriptsize
\setlength{\tabcolsep}{1.8pt}
\begin{tabular}{@{}lrrrrrr@{}}
\toprule
Annotation & $n$ & Pair D & O@3 & T@3 & SegF. & Qwen \\
\midrule
No edit logged & 615 & .858 & .785 & .481 & .766 & .472 \\
Edit logged & 712 & .643 & .574 & .358 & .575 & .328 \\
\bottomrule
\end{tabular}
\caption{Agreement and Dice by workflow across all three experts. A task has an edit logged if any expert recorded a geometry action; each task contributes once. Pair D is raw pairwise expert Dice; O@3 and T@3 are Oracle@3 and Template@3. SegF. is SegFormer-B0; Qwen is Qwen3-VL-32B. Predictions and target-aware oracle masks are fixed before grouping.}
\label{tab:workflow_stress}
\end{table}

\Cref{tab:workflow_stress} shows that tasks with an edit logged have lower expert agreement and lower Dice for all
four methods. Agreement, Oracle@3, and SegFormer are lower in every class;
Template@3 and Qwen are lower in eight of nine. The exception has only one
no-edit symmetric task. Full metrics, confidence intervals, and results by
class and individual expert are in the supplement.

\subsection{Variation Across Annotations}
\label{sec:annotation_variation}

The training masks and test masks were produced by different people, and the
test consensus combines three interpretations. The supplement
compares marginal mask statistics on different image splits. These differences
combine image content, annotator choices, and consensus construction; they do not
isolate an annotator effect.

The training masks are close in area to two experts and smaller than the
third, while containing more connected components. Rescoring representative
methods against each expert and both consensus targets preserves their central
ordering: the greedy oracle measures coverage using the target, while
SegFormer-B0 is the strongest deployable method. Using all proposals, the template exceeds the
best VLM, but their Dice scores are similar using the same 12 proposals (\Cref{tab:matched12}). Training provides supervision; the three-expert test split defines evaluation.

\subsection{Label Conditioning}

On 244 multi-label images, 264 same-image, different-label task pairs have mean
Dice .3655 and IoU .3247. Expert masks therefore depend on the queried label
rather than reducing to one generic foreground mask. Pair-level and
class-template controls are reported in the supplement.

\subsection{Evidence Grounding Baselines}

\begin{table}[t]
\centering
\scriptsize
\setlength{\tabcolsep}{2.0pt}
\begin{tabular}{@{}p{0.37\linewidth}rrrrr@{}}
\toprule
Method & Dice$\uparrow$ & IoU$\uparrow$ & B-F1$\uparrow$ & Over.$\downarrow$ & Hit$\uparrow$ \\
\midrule
Human reference & .867 & .805 & .754 & .113 & .931 \\
Random matched area & .238 & .139 & .037 & .717 & .007 \\
Train class template & .462 & .325 & .169 & .481 & .494 \\
SAM2.1 Oracle@3 & .672 & .605 & .632 & .067 & .661 \\
SAM2.1 Template@3 & .415 & .334 & \textbf{.447} & .510 & .399 \\
U-Net, no label & .496 & .379 & .307 & .458 & .534 \\
U-Net + label & .597 & .479 & .390 & .390 & .661 \\
SegFormer-B0 & \textbf{.664} & \textbf{.576} & .437 & \textbf{.300} & \textbf{.734} \\
Qwen3-VL-32B & .395 & .324 & .440 & .523 & .376 \\
\bottomrule
\end{tabular}
\caption{Evidence grounding on the ambiguity-aware target. Human: held-out
expert versus aligned peer consensus. Proposal rows use the full bank; Qwen
uses 12. Bold marks the best deployable result per column.}
\label{tab:evidence_segmentation}
\end{table}

\begin{figure}[t]
\centering
\includegraphics[width=\linewidth]{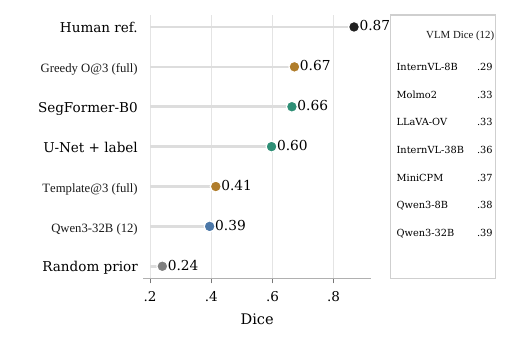}
\caption{Evidence-grounding performance. Representative Dice scores show the
gaps among priors, VLM and template selection, supervised segmentation,
proposal coverage, and human agreement. Full-bank proposal methods and
12-candidate VLMs are identified; matched controls are in \Cref{tab:matched12}. The inset compares seven VLMs using
abbreviated names; full names and metrics are in the supplement.}
\label{fig:evidence_grounding_ladder}
\end{figure}

\Cref{tab:evidence_segmentation,fig:evidence_grounding_ladder} show the gap between coverage and selection:
Oracle@3 reaches .672 Dice,
while Template@3 reaches .415. Label conditioning improves U-Net Dice from
.496 to .597. SegFormer-B0 nearly matches oracle overlap but remains well below
human reliability. On the strict target, Oracle@3, SegFormer-B0, and
Qwen3-VL-32B reach .679, .662, and .399 Dice, preserving the same conclusion.

\paragraph{Modality diagnostic.}
The exploratory segment-versus-line ratings have modest agreement and are
strongly class-structured. A class prior estimated from the training ratings
reaches .7204 accuracy, while the strongest VLM reaches .4348. This diagnostic
characterizes the auxiliary metadata; full VLM results are in the supplement.

\subsection{VLM Proposal Selection}

For each task, selected proposal IDs are joined and evaluated as a mask.
The seven fixed VLMs range from .2910 to .3946 Dice. \Cref{tab:matched12} compares the unchanged oracles and template computed
from training masks using only the proposals originally shown to each VLM. Of 1,327 tasks, 67 have fewer than 12 proposals. Qwen3-VL-32B
and Template@3 now reach .3946 and .4016; the paired Dice difference is
$-.0070$ (95\% task-bootstrap CI $[-.0216,.0075]$). Thus the full-bank gap
partly reflects availability, while both selectors remain well below the exact
up-to-three Dice ceiling of .6448. The full-bank greedy reference reaches
.6718 versus .6400 on 12; greedy selection is not a guaranteed ceiling.
Strict results, all seven VLMs, per-class scores, and uncertainty are in the
supplement.

\begin{table}[t]
\centering\scriptsize
\setlength{\tabcolsep}{4pt}
\begin{tabular}{lrrr}
\toprule
Same 12 candidates & Dice & IoU & B-F1 \\
\midrule
Template@3 & .402 & .320 & .441 \\
Qwen3-VL-32B & .395 & .324 & .440 \\
Greedy Oracle@3 & .640 & .574 & .596 \\
Exact, exactly 3 & .640 & .575 & .596 \\
Exact, up to 3 & .645 & .582 & .610 \\
\bottomrule
\end{tabular}
\caption{Matched-budget selection on the original VLM-visible candidates.
All rows use the same ordered bank, with fewer candidates when unavailable.
Exact oracles optimize Dice using the target; they are not predictors.}
\label{tab:matched12}
\end{table}

Supplementary examples show supervised successes, errors in selecting
proposals, and targets that proposals fail to cover.

\subsection{Auxiliary Recognition Analyses}

\paragraph{Post-hoc classifier explanations.}

This auxiliary diagnostic asks whether successful image-level recognition
localizes human evidence. Its separately trained classifier is distinct from
the supervised region predictors. We train a ResNet-50 on the KU-PCP training labels
and evaluate Grad-CAM for the queried class. On KU-PCP test labels, the
classifier reaches .9012 mean average precision (mAP), .8215 macro-F1, and
.9541 label accuracy at a .5 threshold. We also evaluate DINO ViT
patch-similarity maps. The supplement reports area-matched thresholds and the strongest result
from a fixed area sweep for both heatmap families.

Classifier heatmaps, especially Grad-CAM, can overlap expert masks. Low
boundary agreement distinguishes recognizing labels from localizing the
evidence people select.

\paragraph{Counterfactual deletion.}

Counterfactual deletion tests whether a mask identifies regions used by the
composition classifier. We either blur the masked region or replace it with
the image mean, then measure the target-class logit drop and the fraction of
predictions that fall below threshold. These perturbations are applied to the
classifier described above.

\begin{table}[t]
\centering
\scriptsize
\setlength{\tabcolsep}{2.5pt}
\begin{tabular*}{\columnwidth}{@{\extracolsep{\fill}}lccccc}
\toprule
Mask & Blur $\Delta$ & Fill $\Delta$ & Blur flip & Fill flip & Dice \\
\midrule
Human & 1.02 & 2.03 & .10 & .21 & 1.00 \\
Random area & .16 & .38 & .02 & .03 & .28 \\
Full image & 2.31 & 4.08 & .26 & .76 & .41 \\
U-Net + label & 1.16 & 2.80 & .12 & .28 & .60 \\
SAM O@3 & .55 & 1.31 & .06 & .14 & .67 \\
SAM T@3 & .71 & 1.75 & .07 & .19 & .41 \\
Qwen3-32B & .62 & 1.58 & .06 & .17 & .39 \\
\bottomrule
\end{tabular*}
\caption{Counterfactual deletion utility. Human masks have larger effects than
random masks of equal area. Full-image and overcomplete masks can cause
larger drops, so deletion complements rather than replaces human alignment.
O@3 denotes Oracle@3 and T@3 denotes Template@3.}
\label{tab:counterfactual_deletion}
\end{table}

In \Cref{tab:counterfactual_deletion}, human masks produce much larger drops than random masks of the same area,
supporting their functional relevance. Yet full-image and U-Net masks cause
larger drops despite weaker human alignment, and Template@3 exceeds the
higher-Dice Oracle@3. Regions affecting the classifier and those selected by people are therefore
related but not interchangeable.

\section{Discussion}
\label{sec:discussion}

KU-PCPE turns a composition category into an image-specific grounding task.
Low overlap between different labels on the same image supports a target that
depends on the queried composition. The triangle comparison relates geometric
structure to the objects, surfaces, and negative space in human region masks.

The results separate proposal coverage, semantic selection, and learning.
Oracle@3 reaches .672 Dice, showing that candidate regions often cover the
target, while the full-bank template selector and strongest VLM reach .415
and .395. On the same 12 candidates, template and VLM Dice are similar
(.402 and .395), below the exact up-to-three ceiling (.645).
Supplying the queried label improves U-Net from .496 to .597, and SegFormer
reaches .664, close to proposal-oracle overlap but below expert reliability at
.867. These gaps place label-conditioned selection and precise region assembly
at the center of the task. PICD~\cite{zhao2025picd} found that its best
specialized model distinguished compositions more accurately than the tested
VLMs. Here, supervised segmenters also align more closely with human evidence
than VLM selectors. The gap extends from composition discrimination to
identifying the visual structure that explains a label. Classifier explanations and deletion further show
that recognition-sensitive pixels can differ from the evidence people select.

Expert disagreement is structured by class: rule of thirds, center, and
triangle show strong agreement, while horizontal, diagonal, curved, and pattern
admit more alternative supports. Method ordering remains stable under strict
and ambiguity-aware targets. Individual masks and consensus targets, formed with or without complement
alignment, support studies of uncertainty and alternative valid evidence.

\paragraph{Scope and limitations.}
Candidate regions were used during annotation. Tasks with geometry actions recorded by any of the three experts have lower
agreement and lower model scores. The workflow comparison describes an
association with task difficulty; it does not isolate candidate quality or the
causal effect of editing. All experts used
candidate assistance, so shared candidate bias remains unresolved; a
proposal-free independent audit has not been performed. The training split reflects one annotator's style
and serves as learning supervision; final evaluation uses the three-expert test
split.

KU-PCPE grounds the positive labels inherited from KU-PCP. Extending the
benchmark to negative queries would require a separate target definition: the
absence of symmetry or a triangle may arise without a localized
counter-evidence region. A complete system could combine label prediction with
KU-PCPE grounding for positive decisions. The segment-versus-line ratings are
exploratory, and their modest agreement supports class-level analyses while
instance-level modality remains uncertain. The benchmark covers KU-PCP's nine
labels and photographic images; new domains would require their own definitions
and agreement studies.

\section{Conclusion}

Composition is inherently visual: its meaning in an image depends on the
regions and relations through which people perceive it. KU-PCPE makes those
regions explicit for training and evaluation across nine classes, retaining
shared judgments and alternative interpretations through individual masks and
two consensus views. Specialized segmenters agree more closely with people
than the tested VLM selectors, but remain below agreement among experts.
The dataset therefore supports comparisons between the visual evidence
selected by people and models, providing a step toward understanding how
machine learning perceives stylistic attributes and how that perception
relates to our own.

\clearpage
{
    \small
    \bibliographystyle{ieeenat_fullname}
    \bibliography{main}
}

\section*{Societal Impact}
KU-PCPE masks describe photographic structure rather than identity or sensitive
attributes. Original participant identifiers and calendar timestamps are excluded.
The complete export includes pseudonymous tool counts and relative timelines
grouped into five-second bins. Source photographs are not redistributed; users
reconstruct the annotations against an independently obtained KU-PCP copy under
its original terms. The annotation study involved adult annotators. Permission covers continued processing and pseudonymized release.

\section*{Acknowledgments}
This work was supported by the Austrian Science Fund (FWF), project DFH 37-N,
``Visual Heritage: Visual Analytics and Computer Vision Meet Cultural Heritage''.

\end{document}


\maketitle

\section{Annotation Protocol and Annotators}
\label{supp:protocol}

Each KU-PCPE task showed one photograph and one positive KU-PCP composition
label. The prompt asked annotators to select the segments that explain the
queried composition. The same image could appear in several tasks when KU-PCP
assigned it several labels. These tasks were completed separately.

\begin{figure}[t]
\centering
\begin{minipage}[t]{0.49\textwidth}
\centering
\includegraphics[width=\linewidth]{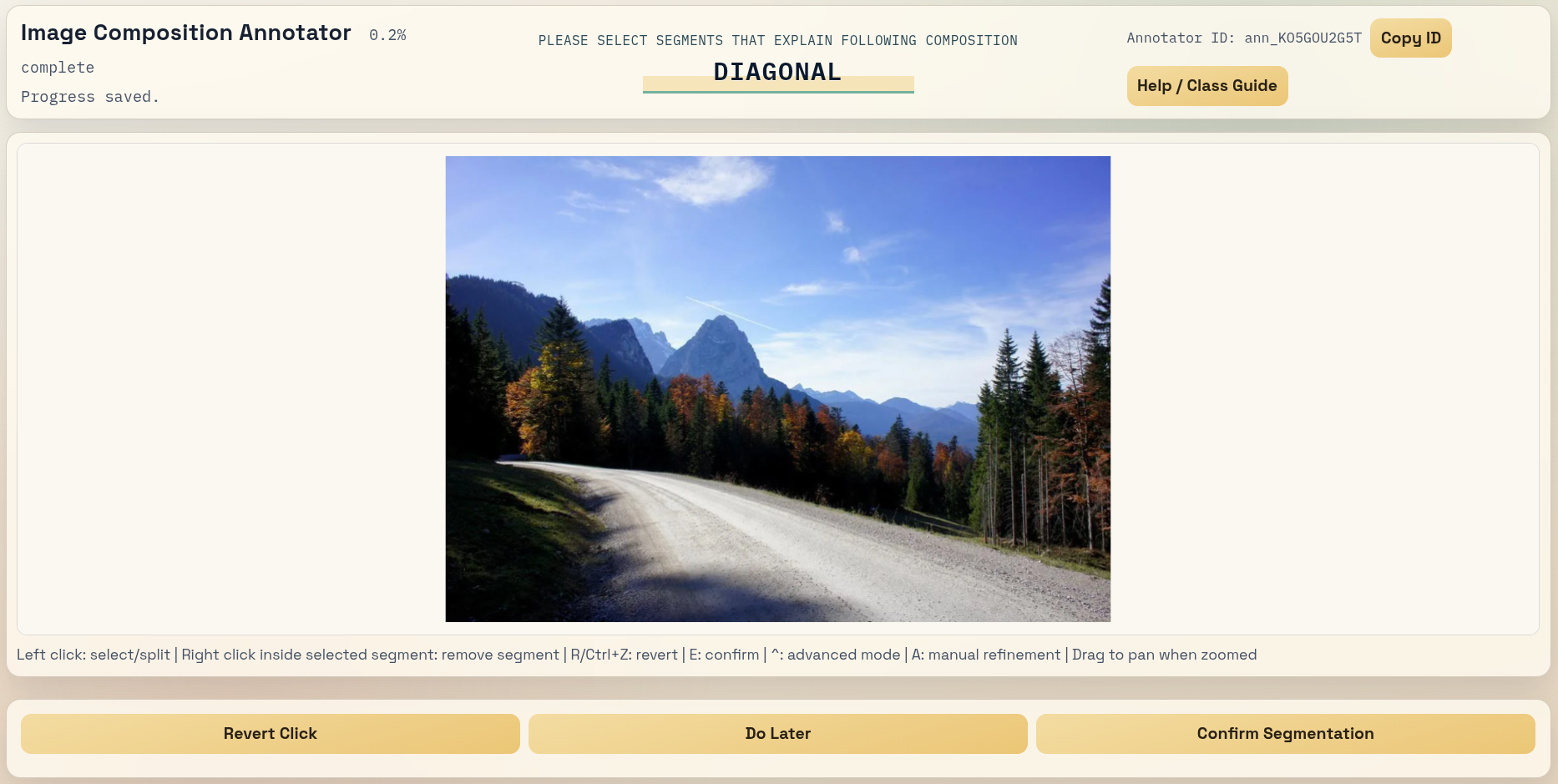}\\[-0.35em]
\small (a) Region selection and mask confirmation.
\end{minipage}\hfill
\begin{minipage}[t]{0.49\textwidth}
\centering
\includegraphics[width=\linewidth]{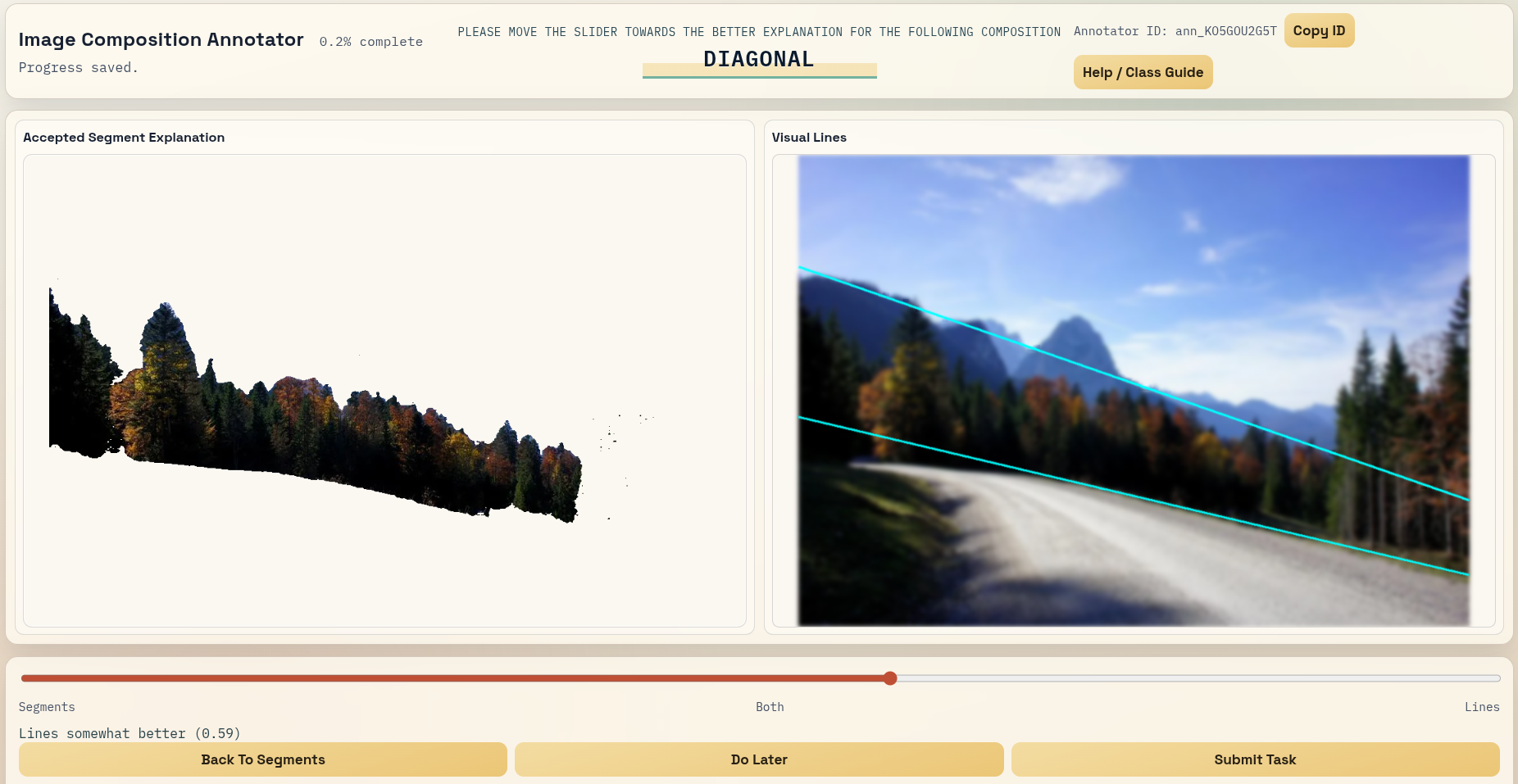}\\[-0.35em]
\small (b) Segment-versus-line rating.
\end{minipage}
\caption{The two annotation states. Candidate regions support efficient mask
editing; the semantic-line overlay is a comparison aid rather than a target.}
\label{fig:supp_interface}
\end{figure}

The interface initially displayed the image and the queried label. Clicking
in the image revealed precomputed candidate regions; a preview showed the
current mask (\Cref{fig:supp_interface}a). Annotators could select several candidates, add a
new region, split or remove a region, refine its boundary, undo changes, and
confirm the final union. After confirmation, they could compare the region
mask with an automatically detected semantic-line overlay
(\Cref{fig:supp_interface}b) and record a soft
segment-versus-line rating. This rating did not alter the binary mask.

The three test experts worked independently and did not see one another's
masks. All had formal education in arts or design and more than five years of
professional experience in art, design, photography, or professional visual
practice. A separate annotator completed the training split. The train and
test annotations used the same image-label task, interface, and class guide.
The training masks provide model supervision; they are not used to estimate
expert agreement.

All scheduled tasks concerned labels already present in KU-PCP. They were not
turned into negative examples when an image was difficult. The released set
contains only completed tasks, and no released mask is empty. Before consensus
construction, exports were checked for complete task coverage, readable masks,
correct image dimensions, and nonempty selections. Flagged masks were reviewed
and corrected before the final test set was formed.

The preserved annotation caches identify SAM2 Hiera-L with configuration
\texttt{configs/sam2/sam2\_hiera\_l.yaml} and checkpoint
\texttt{sam2\_hiera\_large\_072824.pt}. Cached candidates retain their
original ID and score ordering and allow exact replay. The historical
generation thresholds were not stored in those records. The benchmark
uses the separately specified SAM2.1 bank in Supplement~H.1.

\section{Class Guide and Binary Mask Interpretation}
\label{supp:guide}

\subsection{Guide Used During Collection}

The definitions and ``what to look for'' cues in
\Cref{tab:supp_collection_guide} reproduce the guide
given to annotators, with only the layout condensed. \Cref{supp:examples} provides separate examples selected from the benchmark.

\small
\begin{longtable}{@{}p{0.17\textwidth}p{0.35\textwidth}p{0.41\textwidth}@{}}
\caption{Composition definitions and visual cues given to annotators.}
\label{tab:supp_collection_guide}\\
\toprule
Class & Definition & What to look for \\
\midrule
\endfirsthead
\toprule
Class & Definition & What to look for \\
\midrule
\endhead
Rule-of-Thirds (RoT) & Main subject placed along a rule-of-thirds location,
especially near a vertical trisecting line rather than dead center. &
Off-center subject placement; alignment near thirds-grid intersections or
vertical thirds lines; balance created by asymmetrical composition. \\
Vertical & Strong repeated or dominant vertical elements emphasizing height or
upright structure. & Dominant upright lines or repeated vertical rhythm;
height-emphasizing composition; vertical structure outweighs horizontal flow. \\
Horizontal & Strong horizontal structure or leading line, often horizon-like
or calm lateral division. & Dominant left-to-right directional structure;
horizon-like or layered horizontal bands; lateral organization more important
than vertical cues. \\
Diagonal & Dominant slanted line or directional structure that creates
diagonal flow. & Strong angled composition direction; motion or tension
created by slanted alignment; diagonal path that guides gaze. \\
Curved & Dominant curved flow or path guiding the eye. & Arc-like or S-shaped
compositional path; curve-driven eye guidance across the frame; smooth
curvature dominates structural reading. \\
Triangle & Triangular or implied triangular arrangement, including
vanishing-line based triangles. & Three-point or wedge-like structural
arrangement; implied triangular balance in key regions; triangle formed by
lines, edges, or grouped elements. \\
Center & Main subject placed at or very near the image center. & Dominant
visual mass around image midpoint; supporting structure reinforcing central
placement; minimal directional pull away from center. \\
Symmetric & Mirrored structure around a dominant axis; here this is mainly
horizontal symmetry. & Clear mirrored balance around one axis; matching
structural regions on both sides; axis-driven composition dominates
interpretation. \\
Pattern & Repetition of similar shapes or structures creating a patterned
visual effect. & Repeating visual units; regularity or rhythmic repetition;
pattern itself acts as composition driver. \\
\bottomrule
\end{longtable}
\normalsize

\subsection{Binary Mask Interpretation}

The guide identifies the composition to look for. The binary annotation is the
union of visible image regions selected as explaining that composition. It
does not impose one object granularity: a valid union may contain an object,
an object part, several disconnected supports, background, or negative space.
Words such as line, path, and axis describe the relation being explained; the
target is the visible carrier or partition, not a synthetic guideline. When
the initial candidates did not express the intended support, the interface
allowed additions, splits, removals, and manual refinement.
\Cref{tab:supp_mask_interpretation} explains how the same type of mask can
represent these different compositions and where alternative selections arise.

\begin{table}[t]
\centering
\small
\begin{tabularx}{\textwidth}{@{}p{0.18\textwidth}YY@{}}
\toprule
Class & How the mask can express the composition & Common ambiguity \\
\midrule
Center & Centrally placed subject, part, or supporting central region. & Main
subject alone versus a larger central structure. \\
Rule of thirds & Off-center subject or region whose placement creates the
thirds relation. & Subject boundary versus surrounding context; left versus
right placement. \\
Horizontal & Regions or elements forming a lateral division or arrangement. &
Upper versus lower side of a complementary partition. \\
Vertical & Dominant or repeated upright supports. & Complete structures versus
their most salient vertical parts. \\
Diagonal & One elongated carrier or several aligned elements creating slanted
flow. & One side of a diagonal partition versus the other. \\
Symmetric & Corresponding reflected regions and, where relevant, their central
support. & One side, both sides, or the region joining them. \\
Curved & One carrier or several elements jointly forming a perceptual curve. &
The curve carrier versus a broader enclosing region. \\
Triangle & Objects, parts, surfaces, or grouped elements forming an explicit
or implied triangle. & Triangle-forming elements versus the enclosed region. \\
Pattern & Several repeated instances whose regularity drives the composition. &
A representative subset versus the full repeated field. \\
\bottomrule
\end{tabularx}
\caption{How the common region representation applies to the nine classes.
These descriptions clarify the mask representation and were not additional
collection instructions.}
\label{tab:supp_mask_interpretation}
\end{table}

\section{Examples Across All Classes}
\label{supp:examples}

\begin{figure}[p]
\centering
\includegraphics[width=\textwidth]{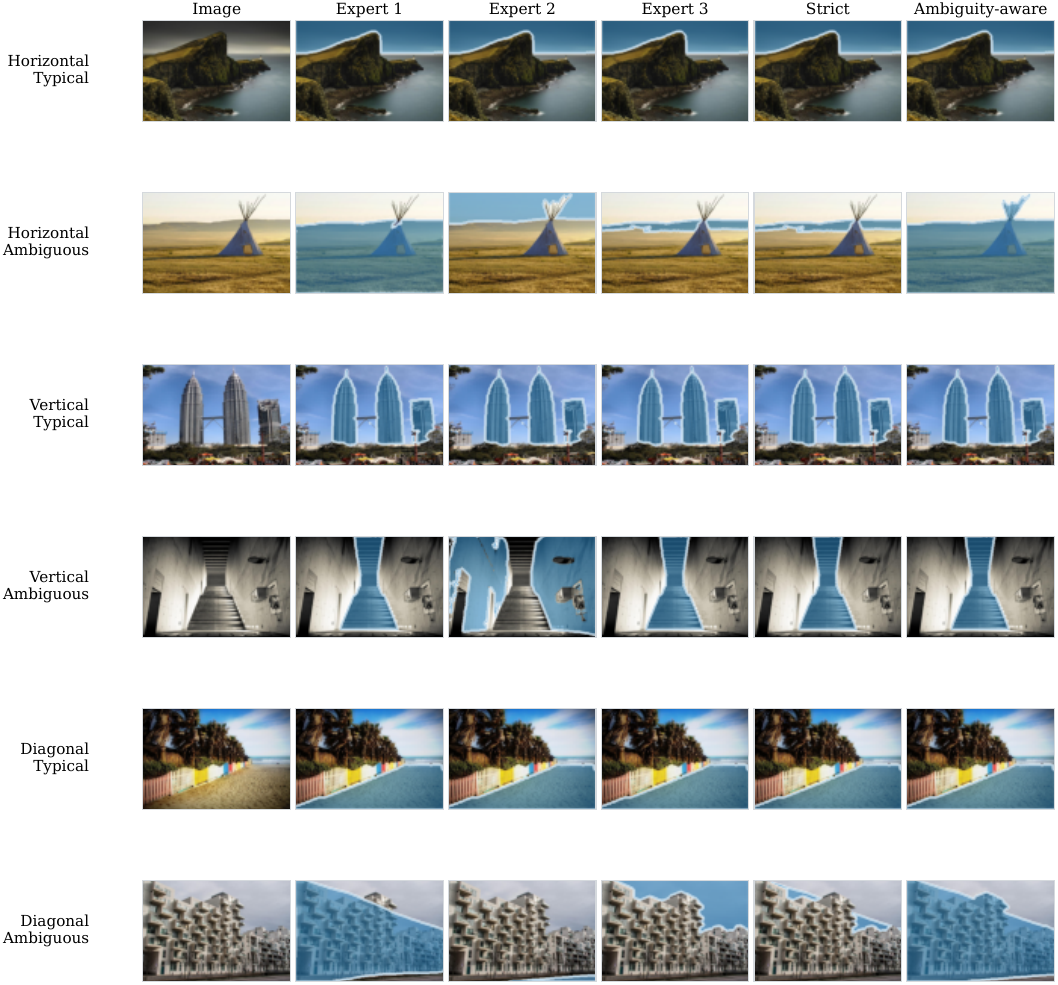}
\caption{Typical and ambiguous examples for horizontal, vertical, and
diagonal. The ambiguous rows show complementary partitions or disagreement
about which visible carrier best expresses the direction.}
\label{fig:supp_examples_1}
\end{figure}

\begin{figure}[p]
\centering
\includegraphics[width=\textwidth]{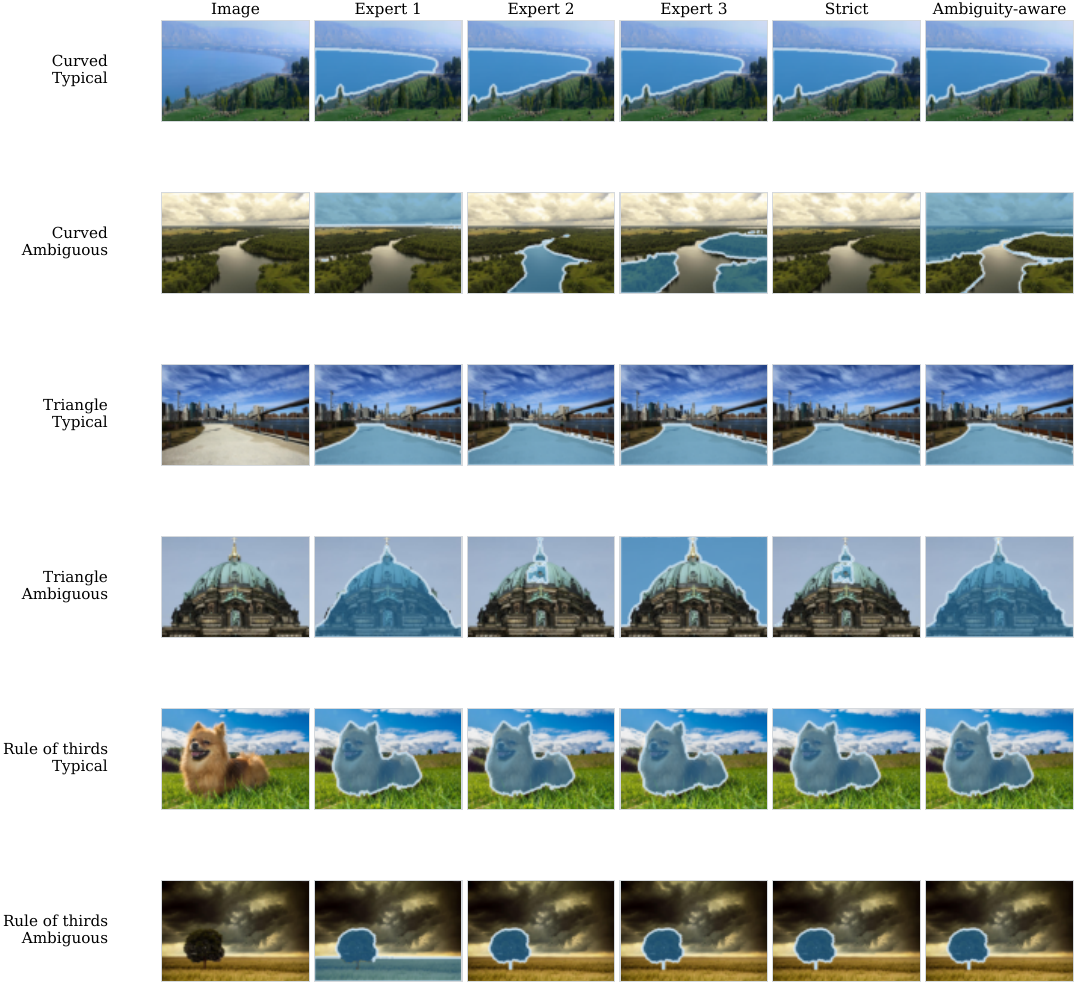}
\caption{Typical and ambiguous examples for curved, triangle, and rule of
thirds. Ambiguity concerns the curve carrier, triangle-forming regions versus
their interior, and subject extent or supporting context.}
\label{fig:supp_examples_2}
\end{figure}

\begin{figure}[p]
\centering
\includegraphics[width=\textwidth]{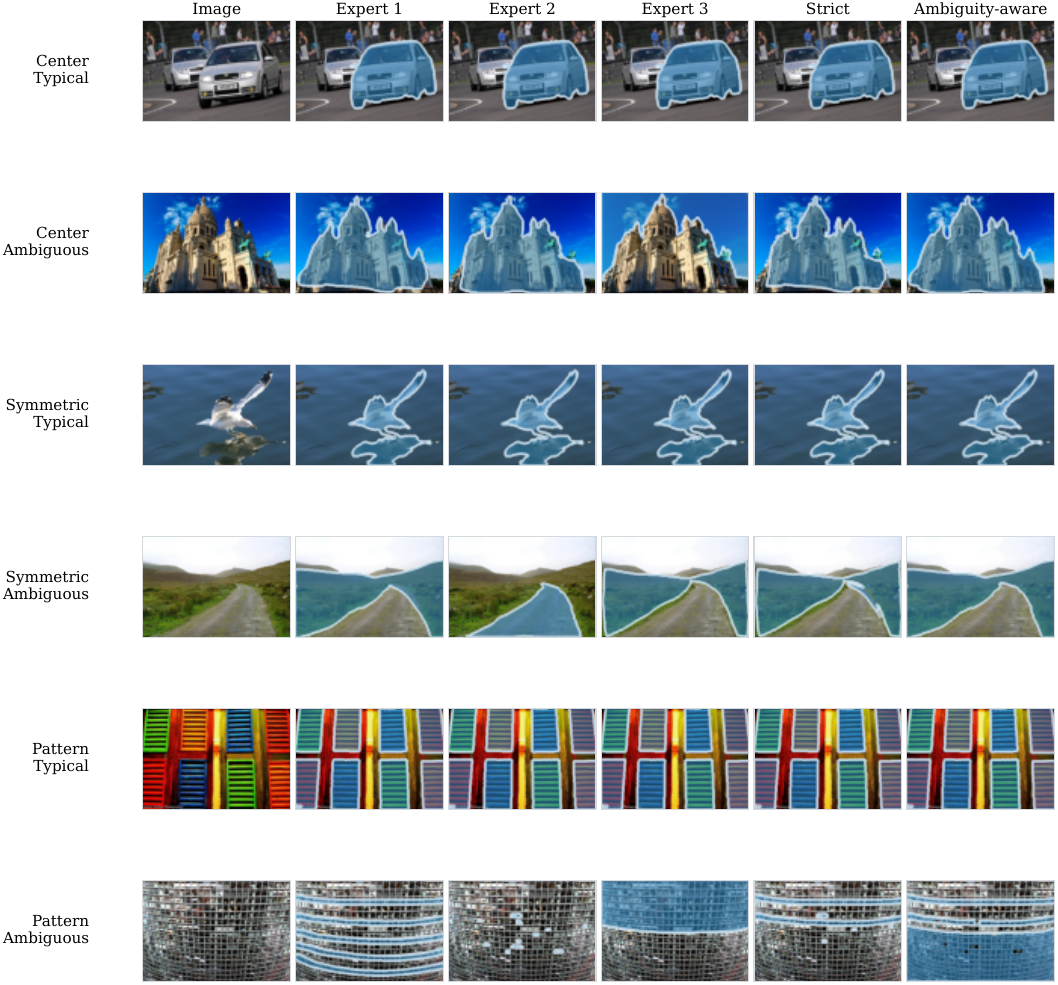}
\caption{Typical and ambiguous examples for center, symmetric, and pattern.
The difficult rows show variation in subject extent, paired versus surrounding
support, and how many repeated instances to include.}
\label{fig:supp_examples_3}
\end{figure}

\Cref{fig:supp_examples_1,fig:supp_examples_2,fig:supp_examples_3} show one
high-agreement task and one ambiguous task for every class. These are selected
from the released expert test set and are benchmark examples, not reconstructed
collection-time guide images. All masks use the same blue overlay and boundary
style as the main paper.
\Cref{tab:supp_example_issues} identifies what the experts interpret
differently in each ambiguous example, from opposite sides of a boundary to
different extents of the same subject.

\begin{table}[p]
\centering
\small
\begin{tabularx}{\textwidth}{@{}p{0.20\textwidth}Y@{}}
\toprule
Class & Representational issue illustrated by the ambiguous row \\
\midrule
Horizontal & Complementary sides of a lateral partition and an intersecting
subject. \\
Vertical & Central upright passage versus surrounding upright walls. \\
Diagonal & Opposite image regions separated by the same slanted boundary. \\
Curved & Water, land, or sky regions that jointly make the curved path visible. \\
Triangle & Triangle-forming architectural regions versus the broader enclosed
shape. \\
Rule of thirds & Main subject alone versus subject with nearby supporting
context. \\
Center & Central architectural mass versus a smaller central portion. \\
Symmetric & Paired reflected regions versus the central path and its
surroundings. \\
Pattern & A subset of repeated units versus the larger repeated field. \\
\bottomrule
\end{tabularx}
\caption{The ambiguity represented in each difficult example.}
\label{tab:supp_example_issues}
\end{table}
\clearpage

\section{Consensus and Agreement}
\label{supp:consensus}

The descriptive agreement table (\Cref{tab:supp_per_class_agreement}) jointly chooses the complement assignments
of experts 2 and 3 to maximize mean pairwise Dice, keeping expert 1 fixed.
Its aligned kappa, majority Dice, and flip rate use that descriptive convention.
The benchmark targets use the reference-based rule below; these two
constructions differ on 48 of the 1,327 tasks.

Let $M_1, M_2, M_3$ be the three expert masks for one image-label task. The
strict target is their pixelwise majority. The ambiguity-aware target uses
$M_1$ as a reference orientation. For $j\in\{2,3\}$, it replaces $M_j$ by its
complement only when $\operatorname{Dice}(\neg M_j,M_1)>
\operatorname{Dice}(M_j,M_1)$, then takes the pixelwise majority. The rule is
deterministic and local to each task. It handles complementary choices around
the same partition; it does not assume complement equivalence for every mask.

In pseudocode:
\begin{quote}\small\ttfamily
strict = majority(M1, M2, M3)\\
aligned = [M1]\\
for M in [M2, M3]:\\
\hspace*{1em}if Dice(not M, M1) $>$ Dice(M, M1): append not M\\
\hspace*{1em}else: append M\\
ambiguity-aware = majority(aligned)
\end{quote}

\begin{figure}[t]
\centering
\includegraphics[width=\textwidth]{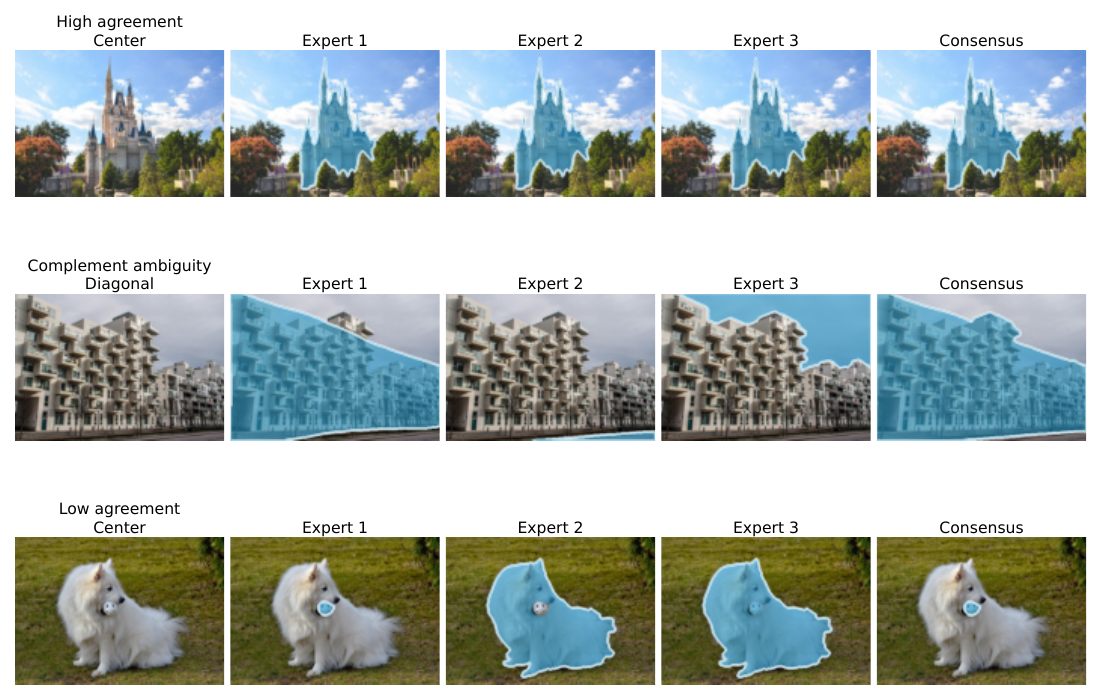}
\caption{High agreement, complementary ambiguity, and low agreement. The
middle row illustrates how opposite sides of one partition can align in the
ambiguity-aware view.}
\label{fig:supp_alignment_example}
\end{figure}

For held-out expert evaluation, the other two masks are aligned by the same
rule and their pixelwise mean is thresholded at .5. With two binary masks this
is their union. The held-out mask is also complemented only if doing so
improves Dice to the peer target. Metrics are averaged over the three held-out
choices within each task. Confidence intervals then resample the 1,327 tasks,
so the three experts are not treated as independent examples.
\Cref{tab:supp_human_reference} summarizes this reference; the high overlap
indicates substantial agreement after accounting for complementary choices.
\Cref{fig:supp_alignment_example} distinguishes those choices from
disagreement about which regions matter. Rule of thirds, center, and triangle
have particularly high raw agreement (\Cref{tab:supp_per_class_agreement}).

\begin{table}[t]
\centering
\small
\begin{tabular}{lcc}
\toprule
Metric & Mean & 95\% task-bootstrap CI \\
\midrule
Dice & .8666 & [.8577, .8750] \\
IoU & .8054 & [.7939, .8167] \\
Boundary F1 & .7541 & [.7411, .7669] \\
Overcoverage & .1131 & [.1058, .1203] \\
Hit@0.5 & .9312 & [.9209, .9410] \\
\bottomrule
\end{tabular}
\caption{Held-out expert agreement with the aligned consensus of the other two
experts.}
\label{tab:supp_human_reference}
\end{table}

\begin{table}[t]
\centering
\scriptsize
\setlength{\tabcolsep}{3.2pt}
\begin{tabular}{lrrrrrrrr}
\toprule
Class & $n$ & Area & Pair D & Pair I & B-F1 & Maj. D & $\kappa$ & Flip \\
\midrule
Horizontal & 211 & .3882 & .6371 & .6241 & .7320 & .9539 & .8281 & .5118 \\
Vertical & 66 & .2426 & .6557 & .5599 & .6619 & .8362 & .5782 & .1515 \\
Diagonal & 189 & .3741 & .5973 & .5566 & .6311 & .8805 & .5433 & .4656 \\
Curved & 92 & .2961 & .6105 & .5593 & .6103 & .8556 & .5240 & .3696 \\
Triangle & 171 & .3126 & .8332 & .7860 & .7611 & .9351 & .7796 & .1170 \\
Rule of thirds & 125 & .0939 & .9538 & .9326 & .9473 & .9782 & .9426 & .0080 \\
Center & 311 & .2051 & .8427 & .8049 & .8008 & .9235 & .8014 & .0354 \\
Symmetric & 99 & .4114 & .8239 & .7419 & .5253 & .9165 & .7152 & .0505 \\
Pattern & 63 & .4794 & .5324 & .4224 & .5729 & .8092 & .2510 & .5397 \\
\bottomrule
\end{tabular}
\caption{Agreement by class. Pair D and Pair I are raw pairwise Dice and IoU;
Maj. D is Dice to the aligned majority; Flip is the fraction of tasks with at
least one complement alignment.}
\label{tab:supp_per_class_agreement}
\end{table}

\subsection{Strict Results}

\begin{table}[t]
\centering
\scriptsize
\setlength{\tabcolsep}{2pt}
\resizebox{\textwidth}{!}{%
\begin{tabular}{lccccc}
\toprule
Method & Dice & IoU & B-F1 & Overcoverage & Hit@0.5 \\
\midrule
Random matched area & .2250 [.2191, .2310] & .1312 [.1273, .1350] & .0357 [.0319, .0397] & .7463 [.7371, .7556] & .0015 [.0000, .0038] \\
Train class template & .4512 [.4387, .4635] & .3186 [.3085, .3287] & .1621 [.1555, .1689] & .5104 [.4943, .5268] & .4891 [.4627, .5170] \\
SAM2.1 Oracle@3 & .6786 [.6591, .6979] & .6159 [.5951, .6368] & .6486 [.6298, .6681] & .0838 [.0724, .0956] & .6714 [.6458, .6963] \\
SAM2.1 Template@3 & .4131 [.3941, .4318] & .3341 [.3160, .3521] & .4429 [.4268, .4593] & .5319 [.5112, .5527] & .3956 [.3693, .4213] \\
U-Net, no label & .4821 [.4662, .4980] & .3691 [.3545, .3835] & .3029 [.2921, .3138] & .4962 [.4787, .5140] & .5200 [.4928, .5463] \\
U-Net + label & .5870 [.5715, .6024] & .4725 [.4574, .4877] & .3864 [.3742, .3986] & .4191 [.4021, .4361] & .6511 [.6255, .6767] \\
SegFormer-B0 & .6624 [.6441, .6801] & .5767 [.5586, .5947] & .4335 [.4193, .4481] & .3197 [.3017, .3377] & .7362 [.7129, .7596] \\
Qwen3-VL-32B & .3988 [.3792, .4186] & .3291 [.3102, .3480] & .4383 [.4217, .4550] & .5380 [.5160, .5601] & .3828 [.3564, .4092] \\
\bottomrule
\end{tabular}}
\caption{Results against the strict consensus. Entries are means with 95\%
task-bootstrap intervals.}
\label{tab:supp_strict_results}
\end{table}

\subsection{Ambiguity-Aware Results}

\begin{table}[t]
\centering
\scriptsize
\setlength{\tabcolsep}{2pt}
\resizebox{\textwidth}{!}{%
\begin{tabular}{lccccc}
\toprule
Method & Dice & IoU & B-F1 & Overcoverage & Hit@0.5 \\
\midrule
Random matched area & .2375 [.2315, .2436] & .1394 [.1354, .1433] & .0370 [.0332, .0409] & .7173 [.7068, .7275] & .0068 [.0030, .0113] \\
Train class template & .4617 [.4502, .4734] & .3254 [.3157, .3351] & .1691 [.1624, .1760] & .4815 [.4655, .4975] & .4936 [.4665, .5200] \\
SAM2.1 Oracle@3 & .6718 [.6533, .6903] & .6054 [.5846, .6263] & .6317 [.6124, .6505] & .0674 [.0575, .0776] & .6609 [.6353, .6865] \\
SAM2.1 Template@3 & .4147 [.3963, .4337] & .3337 [.3160, .3517] & .4475 [.4317, .4629] & .5104 [.4895, .5310] & .3986 [.3730, .4258] \\
U-Net, no label & .4958 [.4803, .5115] & .3790 [.3650, .3930] & .3071 [.2964, .3183] & .4585 [.4410, .4757] & .5335 [.5079, .5607] \\
U-Net + label & .5970 [.5820, .6120] & .4791 [.4645, .4939] & .3904 [.3788, .4025] & .3903 [.3739, .4070] & .6609 [.6353, .6850] \\
SegFormer-B0 & .6636 [.6461, .6813] & .5762 [.5582, .5937] & .4368 [.4228, .4511] & .3000 [.2826, .3175] & .7340 [.7106, .7573] \\
Qwen3-VL-32B & .3946 [.3752, .4144] & .3243 [.3060, .3434] & .4396 [.4235, .4555] & .5231 [.5013, .5446] & .3760 [.3512, .4024] \\
\bottomrule
\end{tabular}}
\caption{Results against the ambiguity-aware consensus. Entries are means with
95\% task-bootstrap intervals.}
\label{tab:supp_ambiguity_results}
\end{table}

\Cref{tab:supp_strict_results,tab:supp_ambiguity_results} show that changing
the consensus rule preserves the broad ordering of methods, with SegFormer-B0
remaining the strongest deployable segmenter.
Oracle@3 is optimized independently for each target because it is a
target-aware greedy coverage reference rather than a guaranteed optimum. Every deployable prediction is fixed and
only rescored across the two views.
\clearpage

\section{Variation Across Annotations}
\label{supp:variation}

The train and test image splits differ, and the training split has one
annotator while the test split has three experts. The comparisons below
describe the supervision and evaluation targets without trying to separate
image differences from annotator differences. Components are 8-connected
foreground components. Boundary complexity is contour perimeter divided by
the square root of foreground area. Intervals bootstrap image-label tasks.

\begin{table}[t]
\centering
\scriptsize
\setlength{\tabcolsep}{3pt}
\begin{tabular}{lcccc}
\toprule
Source & Area fraction & Components & Boundary complexity & Modality score \\
\midrule
Train annotator & .2436 [.2381, .2490] & 10.27 [9.55, 11.05] & 9.71 [9.50, 9.91] & .1955 [.1880, .2031] \\
Expert 1 & .2462 [.2368, .2555] & 7.96 [6.98, 9.05] & 8.69 [8.40, 9.00] & .3373 [.3234, .3512] \\
Expert 2 & .2520 [.2426, .2615] & 8.61 [7.26, 10.38] & 8.38 [8.12, 8.65] & .2499 [.2355, .2646] \\
Expert 3 & .2863 [.2767, .2960] & 7.92 [7.09, 8.82] & 9.15 [8.85, 9.47] & .4949 [.4793, .5104] \\
Strict consensus & .2537 [.2444, .2632] & 9.05 [8.12, 10.05] & 8.94 [8.65, 9.23] & .3607 [.3480, .3733] \\
Ambiguity-aware & .2827 [.2722, .2928] & 8.77 [7.83, 9.81] & 8.82 [8.54, 9.12] & .3607 [.3477, .3736] \\
\bottomrule
\end{tabular}
\caption{Mask statistics for the training annotator, each test expert, and the
two consensus targets. Values are means with 95\% task-bootstrap intervals.}
\label{tab:supp_target_statistics}
\end{table}

\Cref{tab:supp_target_statistics} shows differences in mask area and
component structure across annotators and targets. Consensus construction
also changes area and component structure, particularly when complementary
partitions are aligned. This is why training masks are treated as supervision
rather than a human-reliability sample.

\begin{table}[t]
\centering
\small
\setlength{\tabcolsep}{5pt}
\begin{tabular}{lrrrrr}
\toprule
Method & Expert 1 & Expert 2 & Expert 3 & Strict & Ambig. \\
\midrule
Random matched area & .2199 & .2220 & .2419 & .2250 & .2375 \\
Train class template & .4433 & .4471 & .4444 & .4512 & .4617 \\
SAM2.1 Oracle@3 & .6742 & .6948 & .6688 & .6786 & .6718 \\
SAM2.1 Template@3 & .3957 & .4242 & .4006 & .4131 & .4147 \\
U-Net, no label & .4658 & .4545 & .4951 & .4821 & .4958 \\
U-Net + label & .5649 & .5770 & .5867 & .5870 & .5970 \\
SegFormer-B0 & .6323 & .6483 & .6418 & .6624 & .6636 \\
Qwen3-VL-32B & .3775 & .4074 & .3986 & .3988 & .3946 \\
\bottomrule
\end{tabular}
\caption{Dice after scoring each method against the three individual experts
and both consensus targets. Oracle@3 is selected separately for each target;
all other predictions are fixed.}
\label{tab:supp_target_rescoring}
\end{table}

\Cref{tab:supp_target_rescoring} shows the same broad ordering for all five targets: SegFormer-B0 is the
strongest deployable segmenter, Qwen3-VL-32B remains below full-bank Template@3, and
Oracle@3 remains a reference-dependent coverage control. The model comparison
is therefore not an artifact of one expert or one consensus construction.
\clearpage

\section{Relationship to Original KU-PCP Elements}
\label{supp:kupcp}

KU-PCP defines nine image-level composition classes and proposes
class-specific automatic schemes for locating dominant geometric elements,
including subject boxes, leading lines, a symmetry axis, triangles, and sky
regions. These outputs have different geometric forms. KU-PCPE uses one
human-selected binary region representation for every positive image-label
task and provides three interpretations on the test split. Its contribution is
therefore not the first localization of any composition cue, but a common human
grounding target suitable for agreement analysis and cross-class evaluation.

\begin{table}[t]
\centering
\small
\begin{tabularx}{\textwidth}{@{}p{0.24\textwidth}YY@{}}
\toprule
Property & KU-PCP & KU-PCPE \\
\midrule
Primary supervision & Image-level composition labels & Image-label region
evidence masks \\
Localized output & Class-specific boxes, lines or axes, triangles, and regions
produced by automatic methods & Binary regions selected by people \\
Representation across classes & Different geometric forms for different
classes & One mask representation for all nine classes \\
Test interpretations & One output for each supported detector & Three expert
masks and two consensus views \\
Intended use & Composition classification and element detection & Conditional
grounding of human-selected evidence \\
\bottomrule
\end{tabularx}
\caption{Relationship between the two representations.}
\label{tab:supp_kupcp_relation}
\end{table}

\Cref{tab:supp_kupcp_relation} contrasts the class-specific geometric outputs
with the regions selected by people. Because boxes, lines, axes, triangles, and regions do not share one natural
mask conversion, a common cross-class overlap score is not defined. We report
the 171 triangle tasks for which a matched detector output can be evaluated
without replacing the original method with another detector.

The KU-PCP triangle output stores two detected sides as four endpoints. We
rasterize the sides at the method's 10-pixel width. ``Line in target'' is the
fraction of side pixels inside the KU-PCPE target; the tolerant version dilates
the target by five pixels at the detector resolution. ``Target covered'' is
the fraction of KU-PCPE pixels occupied by the sides. To compare regions, we
also fill the triangle induced by the two base points and the mean of the two
apex endpoints. This filled region is reported separately from the native
side output.

\begin{table}[t]
\centering
\scriptsize
\setlength{\tabcolsep}{3pt}
\begin{tabular}{lrrrrrrr}
\toprule
Target & $n$ & Line in target & +5 px & Target covered & Line D & Filled D & Filled I \\
\midrule
Ambiguity-aware & 171 & .398 [.354, .442] & .489 [.441, .536] & .036 & .064 & .585 [.544, .624] & .462 [.422, .501] \\
Strict & 171 & .374 [.332, .415] & .465 [.419, .511] & .036 & .065 & .572 [.530, .612] & .451 [.411, .490] \\
\bottomrule
\end{tabular}
\caption{Comparison with KU-PCP triangle detector output. D and I denote Dice
and IoU; brackets give 95\% task-bootstrap intervals where reported.}
\label{tab:supp_triangle_comparison}
\end{table}

\begin{figure}[t]
\centering
\includegraphics[width=\textwidth]{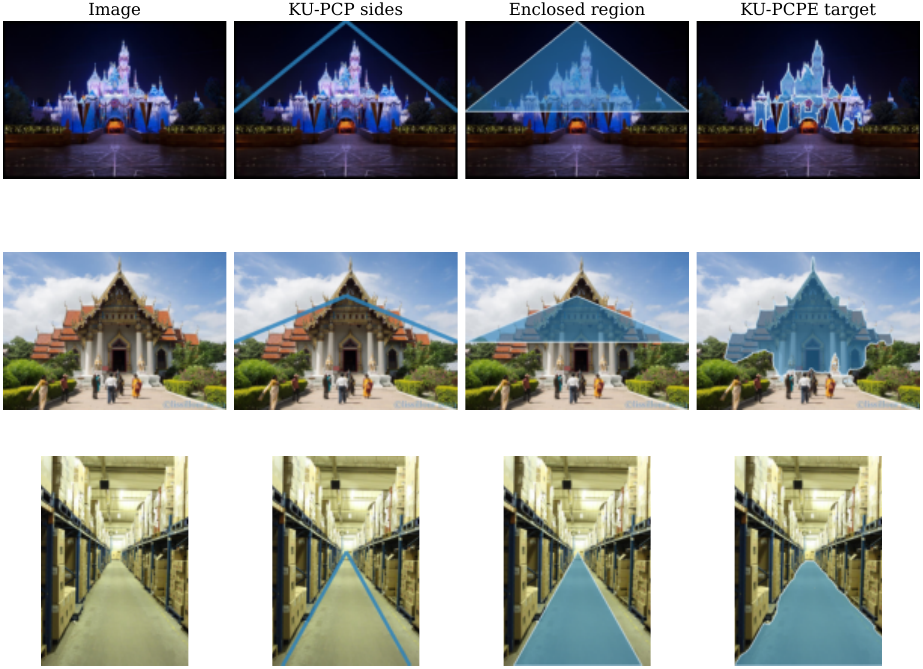}
\caption{Matched triangle examples across the observed overlap range. Columns
show the photograph, KU-PCP's detected sides, the region enclosed by those
sides, and the KU-PCPE ambiguity-aware target.}
\label{fig:supp_triangle_examples}
\end{figure}

In \Cref{tab:supp_triangle_comparison}, the native sides overlap part of the selected human support but cover only
3.6\% of the region target. Filling the triangle raises Dice to .585. This
shows a meaningful relationship to the region organized by the primitive, but
not equivalence between a thin geometric output and the visible regions people
select. \Cref{fig:supp_triangle_examples} illustrates this distinction:
triangle-forming objects and the enclosed image region need not coincide.

\clearpage

\section{Results by Annotation Workflow}
\label{supp:workflow}

The complete export provides retained interaction records for the training
annotator and all three test experts. Process percentages use annotations as
the denominator: 4,093 training and 3,981 test records. Workflow comparisons
instead use 1,327 unique test image-label tasks, each counted once. A task is
\emph{no edit logged} when none of its three experts recorded a geometry
action, and \emph{edit logged} when at least one expert did.

Geometry actions are dynamic region addition, region acceptance, positive or
negative points, applying or undoing manual refinement, splitting, removing a
region, or clearing the current region. Manual tool use includes entering,
applying, exiting, or undoing refinement. These definitions preserve the
original event categories. Event counts and type presence are recoverable from
the released timelines despite their five-second time bins.

The groups describe retained interface records, not complete revision
histories. Discarded drafts are unavailable, and 71 symmetric masks were
replaced after the earlier logged pass. Agreement uses the final three expert
masks; model evaluation uses the final ambiguity-aware consensus. Neither
predictions nor oracle masks are selected anew within a group. The comparison
measures an association with task difficulty, not the effect of editing.

\begin{table}[!htbp]
\centering
\scriptsize
\setlength{\tabcolsep}{2pt}
\resizebox{\textwidth}{!}{%
\begin{tabular}{lrrrrr}
\toprule
Annotation & $n$ & Raw pair D & Comp. pair D & B-F1 & Pair $\kappa$ \\
\midrule
No edit logged & 615 & .8579 [.8384, .8773] & .9132 [.9004, .9257] & .8858 [.8700, .9012] & .8072 [.7792, .8345] \\
Edit logged & 712 & .6434 [.6218, .6652] & .7467 [.7295, .7634] & .5822 [.5631, .6011] & .5292 [.4993, .5589] \\
\bottomrule
\end{tabular}}
\caption{Expert agreement by all-expert workflow group. Brackets give 95\% task-bootstrap intervals. Comp. pair D preserves the descriptive pairwise convention: for each pair in the published expert order, complement the second mask only when Dice improves, then average over the three pairs. It differs from the joint alignment used for majority agreement.}
\label{tab:supp_workflow_agreement}
\end{table}

\begin{table}[!htbp]
\centering
\scriptsize
\setlength{\tabcolsep}{2pt}
\resizebox{\textwidth}{!}{%
\begin{tabular}{lccccc}
\toprule
Method & Dice & IoU & B-F1 & Overcoverage & Hit@0.5 \\
\midrule
Random matched area & .215 [.207, .224] & .125 [.120, .131] & .032 [.026, .037] & .754 [.739, .767] & .008 [.002, .016] \\
Train class template & .485 [.468, .501] & .346 [.331, .360] & .167 [.157, .178] & .478 [.456, .500] & .535 [.496, .574] \\
SAM2.1 Oracle@3 & .785 [.759, .810] & .739 [.710, .767] & .769 [.743, .793] & .025 [.017, .034] & .769 [.735, .802] \\
SAM2.1 Template@3 & .481 [.453, .509] & .397 [.369, .426] & .535 [.513, .557] & .482 [.452, .512] & .465 [.426, .504] \\
U-Net, no label & .558 [.535, .581] & .444 [.422, .466] & .361 [.343, .379] & .431 [.405, .456] & .620 [.582, .659] \\
U-Net + label & .671 [.652, .692] & .556 [.535, .576] & .457 [.439, .475] & .330 [.308, .353] & .767 [.733, .800] \\
SegFormer-B0 & .766 [.743, .788] & .685 [.662, .708] & .524 [.504, .544] & .229 [.206, .252] & .855 [.826, .883] \\
Qwen3-VL-32B & .472 [.443, .502] & .401 [.372, .431] & .535 [.511, .559] & .486 [.455, .518] & .462 [.423, .501] \\
\bottomrule
\end{tabular}}
\caption{Results for the 615 tasks with no geometry action recorded by any expert. Brackets give 95\% task-bootstrap intervals.}
\label{tab:supp_initial_results}
\end{table}

\begin{table}[!htbp]
\centering
\scriptsize
\setlength{\tabcolsep}{2pt}
\resizebox{\textwidth}{!}{%
\begin{tabular}{lccccc}
\toprule
Method & Dice & IoU & B-F1 & Overcoverage & Hit@0.5 \\
\midrule
Random matched area & .257 [.249, .265] & .152 [.146, .157] & .042 [.036, .047] & .686 [.671, .701] & .006 [.001, .011] \\
Train class template & .442 [.426, .457] & .308 [.294, .321] & .171 [.162, .180] & .484 [.461, .507] & .458 [.420, .494] \\
SAM2.1 Oracle@3 & .574 [.548, .599] & .490 [.464, .516] & .513 [.489, .539] & .104 [.088, .121] & .567 [.529, .604] \\
SAM2.1 Template@3 & .358 [.333, .382] & .279 [.256, .301] & .372 [.352, .393] & .535 [.506, .564] & .341 [.306, .375] \\
U-Net, no label & .442 [.423, .461] & .323 [.305, .339] & .261 [.249, .273] & .482 [.459, .505] & .459 [.423, .496] \\
U-Net + label & .533 [.512, .552] & .413 [.393, .432] & .333 [.319, .348] & .442 [.419, .465] & .569 [.532, .605] \\
SegFormer-B0 & .575 [.550, .600] & .482 [.458, .507] & .361 [.343, .379] & .361 [.337, .387] & .629 [.594, .663] \\
Qwen3-VL-32B & .328 [.303, .352] & .258 [.236, .282] & .357 [.337, .378] & .555 [.525, .585] & .302 [.268, .336] \\
\bottomrule
\end{tabular}}
\caption{Results for the 712 tasks with at least one geometry action recorded by at least one expert. Brackets give 95\% task-bootstrap intervals.}
\label{tab:supp_edited_results}
\end{table}

\begin{table}[!htbp]
\centering
\small
\setlength{\tabcolsep}{2.5pt}
\begin{tabular}{llrrrrrr}
\toprule
Annotation & Class & $n$ & Pair D & O@3 & T@3 & SegF. & Qwen \\
\midrule
No edit logged & Center & 203 & .912 & .776 & .548 & .828 & .509 \\
 & Curved & 24 & .822 & .593 & .330 & .552 & .413 \\
 & Diagonal & 61 & .760 & .780 & .477 & .534 & .512 \\
 & Horizontal & 94 & .684 & .917 & .580 & .705 & .613 \\
 & Pattern & 26 & .613 & .319 & .245 & .593 & .215 \\
 & Rule of thirds & 103 & .977 & .915 & .368 & .896 & .393 \\
 & Symmetric & 1 & 1.000 & .995 & .170 & .874 & .170 \\
 & Triangle & 84 & .952 & .686 & .494 & .836 & .350 \\
 & Vertical & 19 & .770 & .854 & .364 & .635 & .664 \\
\midrule
Edit logged & Center & 108 & .713 & .591 & .377 & .565 & .300 \\
 & Curved & 68 & .536 & .526 & .285 & .489 & .234 \\
 & Diagonal & 128 & .520 & .529 & .299 & .414 & .281 \\
 & Horizontal & 117 & .600 & .791 & .558 & .599 & .537 \\
 & Pattern & 37 & .476 & .285 & .170 & .352 & .163 \\
 & Rule of thirds & 22 & .844 & .719 & .203 & .753 & .220 \\
 & Symmetric & 98 & .822 & .458 & .318 & .776 & .303 \\
 & Triangle & 87 & .718 & .507 & .385 & .734 & .254 \\
 & Vertical & 47 & .610 & .712 & .334 & .485 & .496 \\
\bottomrule
\end{tabular}
\caption{Per-class Dice in the two all-expert workflow groups. Abbreviations follow the main paper. Only one symmetric task has no edit logged, so that class cannot support a reliable workflow comparison.}
\label{tab:supp_workflow_per_class}
\end{table}

\Cref{tab:supp_workflow_agreement} shows lower agreement when any expert
recorded an edit. Comparing \Cref{tab:supp_initial_results,tab:supp_edited_results}
shows lower scores for all four methods in that group.
In \Cref{tab:supp_workflow_per_class}, raw agreement, Oracle@3, and SegFormer are lower in the edit group in all
nine classes; Template@3 and Qwen are lower in eight. The exception for the
latter two is symmetric, whose no-edit group has one task. Giving each class
equal weight yields no-edit/edit Dice of .832/.649 for raw agreement,
.759/.569 for Oracle@3, .397/.325 for Template@3, .717/.574 for SegFormer,
and .427/.310 for Qwen. These descriptive averages include the sparse
symmetric group. The direction also holds when each expert's own records
define the groups separately (Table~\ref{tab:supp_workflow_rater_sensitivity}).
\Cref{tab:supp_action_summary} shows that every annotator used geometry tools,
although most annotations had no geometry action recorded. Manual tool use
and other actions vary across annotators.

\begin{table}[!htbp]
\centering
\scriptsize
\setlength{\tabcolsep}{2pt}
\resizebox{\textwidth}{!}{%
\begin{tabular}{lllrrrrr}
\toprule
Rater & Annotation & $n$ & Pair D & O@3 & T@3 & SegF. & Qwen \\
\midrule
rater\_01 & No edit logged & 859 & .793 [.774, .813] & .753 [.731, .775] & .467 [.443, .491] & .709 [.688, .729] & .453 [.428, .478] \\
rater\_01 & Edit logged & 468 & .650 [.624, .676] & .523 [.494, .552] & .319 [.291, .347] & .580 [.550, .610] & .287 [.259, .316] \\
rater\_03 & No edit logged & 770 & .817 [.797, .836] & .767 [.744, .791] & .482 [.456, .507] & .718 [.696, .740] & .462 [.435, .489] \\
rater\_03 & Edit logged & 557 & .640 [.615, .664] & .540 [.512, .568] & .322 [.296, .349] & .588 [.560, .615] & .302 [.275, .329] \\
rater\_04 & No edit logged & 846 & .807 [.788, .825] & .773 [.751, .795] & .484 [.459, .507] & .718 [.697, .739] & .469 [.444, .494] \\
rater\_04 & Edit logged & 481 & .631 [.603, .657] & .493 [.465, .521] & .294 [.267, .320] & .568 [.538, .598] & .264 [.238, .291] \\
\bottomrule
\end{tabular}}
\caption{Sensitivity to the expert whose records define the groups. Each pair of rows partitions all 1,327 tasks using one expert; agreement still uses all three final masks and all models use the same consensus. These are three alternative groupings, not independent cohorts. Brackets give 95\% task-bootstrap intervals.}
\label{tab:supp_workflow_rater_sensitivity}
\end{table}

\begin{table}[!htbp]
\centering
\scriptsize
\setlength{\tabcolsep}{3pt}
\resizebox{\textwidth}{!}{%
\begin{tabular}{llrrrrrrr}
\toprule
Rater & Split & Annotations & No edit & Edit & Manual & Split & Dynamic & Median actions \\
\midrule
rater\_02 & Train & 4,093 & 51.8\% & 48.2\% & 15.1\% & 5.7\% & 31.9\% & 15 \\
rater\_01 & Test & 1,327 & 64.7\% & 35.3\% & 16.5\% & 7.5\% & 27.4\% & 12 \\
rater\_03 & Test & 1,327 & 58.0\% & 42.0\% & 20.0\% & 6.2\% & 30.7\% & 8 \\
rater\_04 & Test & 1,327 & 63.8\% & 36.2\% & 16.1\% & 4.7\% & 28.3\% & 6 \\
\bottomrule
\end{tabular}}
\caption{Interface use for every completed annotator. Percentages refer to annotations, not unique images. No edit/edit use the geometry-action definition above; manual is manual tool use and dynamic is dynamic region addition. Public rater IDs are pseudonyms. The test annotators each completed the same 1,327 tasks.}
\label{tab:supp_action_summary}
\end{table}

\clearpage

\section{Full Benchmark Protocols and Results}
\label{supp:benchmark}

\subsection{Proposal Generation and Selectors}

Benchmark proposals are generated once per test image with SAM2.1 Hiera-L and
the official \texttt{sam2.1\_hiera\_large.pt} checkpoint and
\texttt{sam2.1\_hiera\_l.yaml} configuration. Automatic mask generation uses
32 points per side, 128 points per batch, predicted intersection-over-union
(IoU) threshold .70,
stability threshold .92, stability offset .70, one crop layer, box-NMS
threshold .70, crop-point downscale factor 2, minimum region area 25 pixels,
and mask-to-mask refinement. We retain at most 128 masks, ordered by predicted
IoU and then area. The bank averages 54.48 masks over the 1,073 unique images and 53.77
over the 1,327 image-label tasks, which repeat multi-label images.

\paragraph{Oracle@$k$.}
Starting from an empty mask, the oracle greedily adds the remaining proposal
that gives the greatest Dice for the current union, for at most $k$ additions.
It makes $\min(k,n)$ additions, including an addition that may reduce Dice,
and is recomputed for each target. It is target-aware, not deployable, and
not a guaranteed global optimum.

\paragraph{Template@$k$.}
Training masks are resized and averaged within each class to form a learned
heatmap. Its threshold preserves the median training-mask area. Each test
proposal receives $.7$ times its IoU with this learned template plus $.3$ times
its IoU with a fixed geometric class template, and the top $k$ proposals are
joined.

\paragraph{Other controls.}
Area@$k$ ranks proposals by distance from the median training-mask area for the
queried class. Random@3 samples without replacement and averages ten fixed
draws per task. Random matched area selects a fixed random set of pixels with
the class's median training area. The train class template resizes the learned
heatmap to the test image and thresholds it to that area.

\subsection{Supervised Segmentation}

Training annotations are split 90/10 by image ID with seed 13, so tasks from
one image cannot cross the training and validation sets. Images and masks are
resized to $384\times384$ pixels (bilinear RGB, nearest-neighbor masks).
RGB is scaled to $[0,1]$; SegFormer additionally uses ImageNet mean and
standard-deviation normalization. Training uses a synchronized horizontal flip with probability .5 and AdamW
(weight decay $10^{-4}$) with binary cross-entropy plus soft Dice loss.
Both U-Net variants use 80 epochs, batch size 24, and learning rate
$2\times10^{-4}$. SegFormer-B0 instead
uses 40 epochs, batch size 10, and learning rate $6\times10^{-5}$.
The checkpoint with highest
validation Dice is evaluated, and probabilities are thresholded at .5.

The compact U-Net has three encoder stages, a bottleneck, three decoder stages,
and base width 32. The image-only variant receives RGB. The label-conditioned
variant concatenates nine tiled one-hot label channels to RGB. SegFormer-B0
initializes its encoder from
\texttt{nvidia/segformer-b0-finetuned-ade-512-512}; a learned 32-dimensional
label embedding is tiled over the final encoder feature map before a
three-layer convolutional decoder. Reported results use one fixed seed.

\subsection{Vision-Language Model (VLM) Proposal Selection}

For each image-label task, a VLM receives one rendered image containing up to
the first 12 proposals from the ordered benchmark bank. Each proposal has a
translucent fill, outlined boundary, and integer ID. Decoding is deterministic
and uses greedy generation with at most 64 new tokens. This evaluation takes
place after annotation and is unrelated to how people produced the masks.

The tested checkpoints are Qwen3-VL-32B-Instruct, Qwen3-VL-8B-Instruct,
InternVL3.5-38B-Instruct, InternVL3.5-8B-Instruct, MiniCPM-V-4.5, Molmo2-8B,
and LLaVA-OneVision-Qwen2-7B. The release configuration gives the corresponding
public repository identifiers.

The exact proposal-selection prompt is:
\begin{quote}\small
The image contains numbered SAM2.1 candidate regions. Select up to three
region IDs that best mark the visible evidence supporting the composition
label ``\{label\}''. Return only a JSON list of integers.
\end{quote}

The parser uses the first bracketed substring when present, or otherwise the
full response. It reads integers from left to right, keeps the first three
distinct IDs that refer to rendered candidates, and ignores duplicates and
out-of-range values. If no valid ID remains, or inference fails, the prediction
is an empty mask. Otherwise the accepted proposal masks are joined. The invalid
rate is the fraction of tasks with no accepted ID.

\subsection{Metrics and Confidence Intervals}

For prediction $M$ and target $G$, Dice is
$2|M\cap G|/(|M|+|G|)$ and IoU is $|M\cap G|/|M\cup G|$.
For deployable predictions, Boundary F1 first resizes the binary prediction
and target to a maximum side of 256 using bilinear filtering and a threshold of
128, extracts erosion boundaries, and accepts a match within two pixels.
The historical human leave-one-out reference uses nearest-neighbor resizing
before erosion boundaries. Descriptive pairwise agreement uses Pillow
nearest-neighbor resizing, marks both sides of horizontal and vertical
transitions without an image-frame boundary, and matches within a
$5\times5$ square (Chebyshev distance at most two pixels). These
analysis-specific conventions are preserved. Overcoverage is $|M\setminus G|/|M|$, with value zero for an empty
prediction. Hit@0.5 is the fraction of tasks with Dice at least .5. Confidence
intervals are the 2.5th and 97.5th percentiles of 10,000 bootstrap samples of
image-label tasks, using seed 920. Tasks, rather than pixels or individual
expert masks, are the sampling unit.

\subsection{Post-hoc Explanations and Deletion}

The image-level recognizer is an ImageNet-initialized ResNet-50 trained on
KU-PCP's nine multi-label targets. It uses an 85/15 train/validation split,
224-pixel inputs, color jitter and horizontal flips, binary cross-entropy,
AdamW (learning rate $3\times10^{-4}$, weight decay $10^{-4}$), batch size 64,
and 12 epochs. The checkpoint with best validation mean average precision
(mAP) is retained. Grad-CAM uses the final ResNet stage. The DINO comparison uses
\texttt{vit\_small\_patch16\_224.dino} and cosine similarity between each
patch token and the global token. Heatmaps are normalized per task. The
area-matched mask preserves the expert target's area. The best-area sweep
chooses the strongest target Dice from area fractions .03, .05, .08, .10, .15,
.20, .30, .40, .50, .65, and .80. It is therefore an oracle threshold
diagnostic rather than a deployable setting.

Deletion uses the same ResNet-50. Mean-fill replaces selected pixels with the
image's per-channel mean. Blur replaces them with a Gaussian-blurred image
using an odd kernel of at least 31 pixels and approximately one eighteenth of
the longer image side. A flip is counted when an originally positive target
prediction falls below .5 after perturbation. Full-image deletion is an upper
control, not an evidence mask.

\subsection{Label Conditioning}

\begin{table}[t]
\centering
\small
\begin{tabular}{lr}
\toprule
Diagnostic & Value \\
\midrule
Multi-label images & 244 \\
Different-label task pairs & 264 \\
Expert same-image, different-label Dice & .3655 \\
Expert same-image, different-label IoU & .3247 \\
Train-template same-image, different-label Dice & .3095 \\
Expert/template distance correlation & .4508 \\
\bottomrule
\end{tabular}
\caption{Aggregate label-conditioning diagnostics.}
\label{tab:supp_label_conditioning}
\end{table}

\begin{table}[t]
\centering
\small
\begin{tabular}{lrrrr}
\toprule
Label pair & Pairs & Expert D & Expert I & Template D \\
\midrule
Center / Triangle & 54 & .8106 & .7674 & .5308 \\
Horizontal / Rule of thirds & 40 & .0218 & .0123 & .0000 \\
Diagonal / Horizontal & 37 & .1540 & .1224 & .3202 \\
Curved / Horizontal & 22 & .2301 & .1739 & .1401 \\
Curved / Diagonal & 18 & .3442 & .2846 & .6668 \\
Horizontal / Triangle & 12 & .1591 & .1214 & .2282 \\
Center / Rule of thirds & 10 & .7883 & .7764 & .1091 \\
Center / Symmetric & 9 & .7121 & .5880 & .5696 \\
Triangle / Vertical & 8 & .4837 & .4062 & .1188 \\
\bottomrule
\end{tabular}
\caption{Frequent pairs of different labels assigned to the same image.}
\label{tab:supp_frequent_label_pairs}
\end{table}

\Cref{tab:supp_label_conditioning} shows that masks for different labels on
the same image have limited overlap. The pair-specific values in
\Cref{tab:supp_frequent_label_pairs} show why label conditioning is not equivalent to
generic foreground segmentation. Some labels can share a subject, as in
center and triangle, whereas others direct attention to nearly disjoint
regions in the same photograph.

\subsection{Proposal Budgets and Supervised Models}

\Cref{tab:supp_proposal_budgets} separates how well proposals can cover the
target from how well a method selects them. The supervised comparison in
\Cref{tab:supp_supervised} shows the benefit of supplying the queried label
to U-Net and the stronger overlap achieved by SegFormer-B0.
\Cref{tab:supp_supervised_per_class} shows that performance also varies
substantially across composition classes.

\begin{table}[t]
\centering
\small
\begin{tabular}{lrrrrr}
\toprule
Proposal method & Dice & IoU & B-F1 & Over. & Hit \\
\midrule
Random@3 & .0895 & .0684 & .1224 & .6449 & .0622 \\
Largest & .2491 & .2294 & .3372 & .6950 & .2419 \\
Area@1 & .3571 & .3287 & .3946 & .5352 & .3519 \\
Area@3 & .4133 & .3259 & .4250 & .5676 & .4190 \\
Template@1 & .4099 & .3724 & .4325 & .3901 & .3873 \\
Template@3 & .4147 & .3337 & .4475 & .5104 & .3986 \\
Oracle@1 & .5851 & .5197 & .5480 & .0612 & .5539 \\
Oracle@3 & .6718 & .6054 & .6317 & .0674 & .6609 \\
Oracle@5 & .6966 & .6287 & .6529 & .0755 & .6918 \\
Oracle@10 & .7040 & .6297 & .6444 & .1196 & .7189 \\
\bottomrule
\end{tabular}
\caption{Proposal baselines against the ambiguity-aware target.}
\label{tab:supp_proposal_budgets}
\end{table}

\begin{table}[t]
\centering
\small
\begin{tabular}{lcrrrrr}
\toprule
Method & Label input & Dice & IoU & B-F1 & Over. & Hit \\
\midrule
U-Net & No & .4958 & .3790 & .3071 & .4585 & .5335 \\
U-Net & Yes & .5970 & .4791 & .3904 & .3903 & .6609 \\
SegFormer-B0 & Yes & .6636 & .5762 & .4368 & .3000 & .7340 \\
\bottomrule
\end{tabular}
\caption{Supervised segmentation against the ambiguity-aware target.}
\label{tab:supp_supervised}
\end{table}

\begin{table}[t]
\centering
\small
\begin{tabular}{lrrr}
\toprule
Class & U-Net, no label & U-Net + label & SegFormer-B0 \\
\midrule
Horizontal & .2838 & .4705 & .6459 \\
Vertical & .4465 & .5264 & .5281 \\
Diagonal & .3719 & .4596 & .4531 \\
Curved & .3789 & .4452 & .5052 \\
Triangle & .5889 & .6595 & .7841 \\
Rule of thirds & .6397 & .7840 & .8704 \\
Center & .6133 & .6556 & .7365 \\
Symmetric & .5900 & .8059 & .7770 \\
Pattern & .5337 & .5705 & .4513 \\
\bottomrule
\end{tabular}
\caption{Per-class Dice for supervised models.}
\label{tab:supp_supervised_per_class}
\end{table}

\subsection{VLM Proposal Selection}

\begin{table}[t]
\centering
\small
\setlength{\tabcolsep}{3.5pt}
\begin{tabular}{lrrrrrr}
\toprule
Model & Dice & IoU & B-F1 & Over. & Hit & Invalid \\
\midrule
Qwen3-VL-32B & .3946 & .3243 & .4396 & .5231 & .3760 & .0008 \\
Qwen3-VL-8B & .3770 & .3111 & .4309 & .5622 & .3519 & .0008 \\
MiniCPM-V 4.5 & .3693 & .3081 & .4107 & .5435 & .3429 & .0000 \\
InternVL3.5-38B & .3634 & .3003 & .4114 & .5408 & .3384 & .0211 \\
LLaVA-OneVision-7B & .3312 & .2579 & .3971 & .6608 & .3120 & .0000 \\
Molmo2-8B & .3286 & .2747 & .3909 & .6071 & .3044 & .0000 \\
InternVL3.5-8B & .2910 & .2416 & .3618 & .6124 & .2668 & .0580 \\
\bottomrule
\end{tabular}
\caption{VLM selection of up to three numbered proposals.}
\label{tab:supp_vlm_proposals}
\end{table}

The seven models in \Cref{tab:supp_vlm_proposals} occupy a narrow range below full-bank Template@3. Matched-budget controls are in Section~\ref{sec:matched12}. Larger checkpoints do
not consistently recover the human evidence, and invalid responses are too
rare to explain the gap for most models.

\subsection{Exploratory Modality Ratings}

All 8,074 annotations have a recorded modality score. Ratings are binned as segment for $r<1/3$, mixed for
$1/3\leq r\leq2/3$, and line for $r>2/3$. Among the three test experts, pairwise three-bin agreement is
.5386, Fleiss $\kappa=.2642$, and mean pairwise absolute rating difference is
.2235.
\Cref{tab:supp_modality_distribution} shows that preference for regions or
lines varies by composition class. \Cref{tab:supp_modality_raters} also shows
variation among annotators; the training and test summaries concern different
images and cannot separate annotator effects from image differences.

\begin{table}[t]
\centering
\scriptsize
\setlength{\tabcolsep}{3pt}
\resizebox{\textwidth}{!}{%
\begin{tabular}{lrrrrrr}
\toprule
Class & Train $n$ & Mean $r$ & S / M / L (\%) & Test $n$ & Mean $r$ & S / M / L (\%) \\
\midrule
Horizontal & 758 & .47 & 13 / 83 / 4 & 633 & .55 & 9 / 64 / 27 \\
Vertical & 195 & .11 & 84 / 15 / 1 & 198 & .22 & 70 / 24 / 7 \\
Diagonal & 471 & .44 & 28 / 60 / 12 & 567 & .59 & 12 / 45 / 43 \\
Curved & 235 & .01 & 99 / 1 / 0 & 276 & .14 & 84 / 13 / 3 \\
Triangle & 374 & .34 & 45 / 52 / 4 & 513 & .54 & 22 / 38 / 40 \\
Rule of thirds & 923 & .02 & 98 / 1 / 0 & 375 & .20 & 72 / 22 / 5 \\
Center & 750 & .03 & 98 / 2 / 0 & 933 & .21 & 71 / 24 / 5 \\
Symmetric & 232 & .22 & 72 / 25 / 3 & 297 & .29 & 55 / 37 / 8 \\
Pattern & 155 & .01 & 100 / 0 / 0 & 189 & .17 & 77 / 20 / 3 \\
\bottomrule
\end{tabular}}
\caption{Exploratory modality ratings for both splits. S/M/L denote segment/mixed/line percentages. Training has one rating per task and test has three. The splits contain different images and annotators, so this is an unpaired descriptive comparison. The main figure reports expert-test ratings.}
\label{tab:supp_modality_distribution}
\end{table}

\begin{table}[t]
\centering
\scriptsize
\setlength{\tabcolsep}{3pt}
\resizebox{\textwidth}{!}{%
\begin{tabular}{llrrrrr}
\toprule
Rater & Split & Ratings & Mean $r$ & Segment & Mixed & Line \\
\midrule
rater\_01 & Test & 1327 & .250 & 58.9\% & 33.2\% & 7.9\% \\
rater\_02 & Train & 4093 & .196 & 67.5\% & 29.8\% & 2.8\% \\
rater\_03 & Test & 1327 & .495 & 34.5\% & 27.4\% & 38.1\% \\
rater\_04 & Test & 1327 & .337 & 45.9\% & 44.6\% & 9.5\% \\
\bottomrule
\end{tabular}}
\caption{Modality scores for each completed annotator. These summaries describe recorded scores; they are not evidence of agreement between the training annotator and test experts, who worked on different image splits.}
\label{tab:supp_modality_raters}
\end{table}

\begin{table}[t]
\centering
\small
\begin{tabular}{lrrrr}
\toprule
Modality predictor & Accuracy & Macro-F1 & MAE & Invalid \\
\midrule
Always segment & .4943 & .2205 & .3082 & -- \\
Training-set class prior & .7204 & .5008 & .1398 & -- \\
Molmo2-8B & .4348 & .3522 & .2833 & .0000 \\
Qwen3-VL-32B & .3338 & .2832 & .4668 & .0000 \\
Qwen3-VL-8B & .3157 & .3048 & .4672 & .0000 \\
LLaVA-OneVision-7B & .3263 & .2615 & .4759 & .0000 \\
InternVL3.5-38B & .2976 & .2724 & .4048 & .0023 \\
InternVL3.5-8B & .3090 & .2776 & .4555 & .0000 \\
MiniCPM-V 4.5 & .1959 & .1870 & .5629 & .0000 \\
\bottomrule
\end{tabular}
\caption{VLM diagnostics for exploratory modality metadata. MAE is mean
absolute error.}
\label{tab:supp_vlm_modality}
\end{table}

In \Cref{tab:supp_vlm_modality}, the training-set class prior exceeds every VLM because modality is strongly
class-structured and the expert signal is uncertain. This experiment
characterizes the metadata rather than defining a primary grounding benchmark.

\subsection{Post-hoc Explanations}

The ResNet-50 classifier reaches mAP .9012, macro-F1 .8215, micro-F1 .8258,
label accuracy .9541, and exact-match accuracy .6980.
\Cref{tab:supp_posthoc} shows that classifier heatmaps can overlap the human
masks while agreeing less closely with their boundaries. Recognizing a label
therefore does not ensure that the classifier localizes the regions people select.

\begin{table}[t]
\centering
\small
\begin{tabular}{lrrrrr}
\toprule
Method & Dice & IoU & B-F1 & Over. & Hit \\
\midrule
DINO, area matched & .2102 & .1432 & .1413 & .7898 & .1515 \\
DINO, best area sweep & .3404 & .2314 & .2018 & .7307 & .2524 \\
Grad-CAM, area matched & .5176 & .3793 & .1371 & .4907 & .6074 \\
Grad-CAM, best area sweep & .5833 & .4328 & .1947 & .4927 & .7242 \\
\bottomrule
\end{tabular}
\caption{Classifier-derived explanation masks.}
\label{tab:supp_posthoc}
\end{table}

\subsection{Counterfactual Deletion}

\begin{table}[t]
\centering
\scriptsize
\setlength{\tabcolsep}{3pt}
\begin{tabular}{lrrrrrr}
\toprule
Mask source & Blur drop & Fill drop & Blur flip & Fill flip & Mask D & Area \\
\midrule
Human consensus & 1.0192 & 2.0311 & .0965 & .2095 & 1.0000 & .2827 \\
Random matched area & .1645 & .3754 & .0173 & .0347 & .2827 & .2827 \\
Edge matched area & 1.0981 & 2.2527 & .0942 & .2306 & .4499 & .3246 \\
Full image & 2.3069 & 4.0795 & .2622 & .7611 & .4085 & 1.0000 \\
U-Net + label & 1.1649 & 2.8026 & .1168 & .2833 & .5970 & .3160 \\
SAM2.1 Area@3 & .5675 & 1.5486 & .0603 & .1703 & .4133 & .3294 \\
SAM2.1 Oracle@3 & .5531 & 1.3079 & .0565 & .1417 & .6718 & .1733 \\
SAM2.1 Template@3 & .7084 & 1.7485 & .0701 & .1922 & .4147 & .3899 \\
Qwen3-VL-32B & .6249 & 1.5760 & .0610 & .1726 & .3946 & .3426 \\
\bottomrule
\end{tabular}
\caption{Classifier logit drops and prediction flips after deleting each mask.
Mask D is Dice to the human target.}
\label{tab:supp_deletion}
\end{table}

\Cref{tab:supp_deletion} shows that human masks have much larger effects than random masks with the same area.
However, full-image and overcomplete masks can cause larger drops despite
weaker human alignment, and Template@3 exceeds the higher-Dice Oracle@3.
Deletion therefore supports functional relevance while measuring something
different from agreement with human evidence.
\clearpage

\section{Release and Reconstruction}
\label{supp:release}

KU-PCPE contains positive image-label grounding tasks for nine KU-PCP
composition classes. A binary mask marks visible regions selected as support
for a label already judged applicable. The training split contains one
annotator; the official test split contains three experts represented by pseudonyms and both
strict and ambiguity-aware consensus targets. The masks are subjective
judgments of photographic structure, not identity or sensitive-attribute
labels. Known limitations are candidate-assisted collection, one-annotator
training supervision, positive-only queries, and exploratory modality ratings
with modest agreement.
\Cref{tab:supp_release_files} lists the annotation assets used by the
evaluator: benchmark masks, derived consensus masks, metadata, and validation
files. These supporting assets allow individual expert judgments and both
consensus rules to be evaluated separately.

\begin{table}[t]
\centering
\small
\begin{tabularx}{\textwidth}{@{}p{0.42\textwidth}Y@{}}
\toprule
Path & Contents \\
\midrule
\texttt{annotations/single\_annotator\_train/} & 4,093 records from 3,169
train images and their binary masks \\
\texttt{annotations/expert\_test/expert\_01..03/} & Three pseudonymous bundles
of 1,327 records and masks \\
\texttt{consensus/strict/} & 1,327 strict-majority masks \\
\texttt{consensus/ambiguity\_aware/} & 1,327 ambiguity-aware masks \\
\texttt{consensus/manifest.csv} & Task-to-consensus-mask index \\
\texttt{metadata/preference\_ratings.csv} & Exploratory expert modality
ratings \\
\texttt{metadata/process/} & Scalar workflow labels and aggregate summaries \\
\texttt{split\_manifest.csv} & Split-local image, task, and annotator counts \\
\texttt{schemas/} & Annotation and workflow field definitions \\
\texttt{benchmark\_proposal\_config.json} & Fixed proposal benchmark settings \\
\texttt{file\_manifest.csv}, \texttt{checksums.csv} & Complete inventory and
SHA-256 checksums \\
\texttt{README.md}, \texttt{DATASET\_CARD.md}, \texttt{DATASHEET.md} & Scope,
construction, uses, and limitations \\
\texttt{LICENSE.md}, \texttt{CITATION.cff} & Annotation-layer license and
citation information \\
\texttt{RECONSTRUCTION.md}, \texttt{evaluate\_smoke\_test.py} & KU-PCP join
instructions and integrity test \\
\bottomrule
\end{tabularx}
\caption{Annotation assets for the evaluator. The dataset download described
below contains the complete annotations and metadata for all four annotators.}
\label{tab:supp_release_files}
\end{table}

The dataset download,
\texttt{kupcp\_completed\_public\_20261023\_v1.tar.gz}, contains the complete
annotations and metadata for the training annotator and all three test experts. Every annotation
has a segment-versus-line score, recorded settings, and action counts. It also
includes final and component masks, a pseudonymous participant index, summaries
by annotator, class, and task, and relative tool-event timelines. The separate
reproducibility project includes a script that joins these records to the
fixed per-task benchmark scores and regenerates all-rater workflow tables,
including confidence intervals and per-rater sensitivity results.

To reconstruct the benchmark, users obtain KU-PCP through its original
distribution and set \texttt{KUPCP\_ROOT} to a directory containing
\texttt{train\_img/} and \texttt{test\_img/}. Training records resolve under
the former and expert-test records under the latter. The filename namespaces
remain separate. From the release root, the verification command is:

\begin{quote}\ttfamily\small
KUPCP\_ROOT=/path/to/KU-PCP python -I evaluate\_smoke\_test.py
\end{quote}

The expected final line is:

\begin{quote}\ttfamily\small
SMOKE TEST PASSED: 4093 train tasks; 1327 expert-test tasks; identity
Dice=1.000000
\end{quote}

The integrity test decodes every annotation PNG, verifies binary values and
foreground areas, measures pixel-exact round-trip Dice, and reconstructs both
consensus targets. It also checks every package checksum, mask path, consensus link, task
identity, PNG dimension, split count, and referenced KU-PCP image. It was run
from a fresh copy of the release in an isolated Python process. Source
photographs are not redistributed or relicensed. The KU-PCPE annotation layer
is licensed under CC BY 4.0; users must separately follow KU-PCP's original
terms.

In the complete export, \texttt{tool\_usage\_by\_task.csv} gives tool counts for
each annotation; summaries by annotator and class are supplied separately.
\texttt{tool\_timeline.csv} groups recorded events into five-second bins within
observed clock segments, with time values capped at 30 minutes. Original
participant identifiers, calendar timestamps, coordinates, free text, and raw
event payloads are withheld. These records describe observed interface use,
not exact inference calls or active working time. Discarded drafts and
unretained revision histories are not reconstructed.

\section{Matched 12-Proposal Controls}
\label{sec:matched12}
We replay the exact original first 12 masks in shard order and their 1-based
IDs, without regenerating or re-ranking proposals. All 1,073 image overlays
match the original rendering byte for byte. The 67 tasks (59 distinct images)
with fewer proposals use their available bank. Saved responses from all seven VLMs are evaluated with the parser described
in \Cref{supp:benchmark}.
These are saved-prediction re-evaluations, not new VLM inference runs.

The template scores and training masks are unchanged. Greedy selection makes
$\min(3,n)$ additions even when a later addition reduces Dice, resolving ties
by original candidate order. The exact reference enumerates all subsets with
exactly $\min(3,n)$ masks, and separately all subsets of at most three masks,
including the empty subset. Each oracle optimizes Dice separately for each
target; predictions are never inverted. Other reported metrics evaluate the
Dice-maximizing subset and are not independently optimized ceilings.
Exact integer subset enumeration passed 156 randomized and empty-case checks
against explicit binary unions. Bootstrap intervals resample image-label tasks
10,000 times with seed 920; repeated images are not independent resampling
clusters, so these intervals should not be read as image-cluster uncertainty.

Under both consensus rules, the template and strongest VLM have similar Dice,
while the exact references recover more of the target from the same candidates
(\Cref{tab:matched12_0_0,tab:matched12_0_1,tab:matched12_1_0,tab:matched12_1_1}).

\begin{table}[htbp]\centering\scriptsize
\begin{tabular}{lrrr}\toprule
Method & Dice & IoU & B-F1 \\\midrule
Template@3 & 0.4016 [0.3831, 0.4200] & 0.3204 [0.3030, 0.3382] & 0.4412 [0.4255, 0.4567] \\
Qwen3-VL-32B & 0.3946 [0.3752, 0.4139] & 0.3243 [0.3057, 0.3431] & 0.4396 [0.4232, 0.4556] \\
Greedy Oracle@3 & 0.6400 [0.6202, 0.6595] & 0.5745 [0.5539, 0.5952] & 0.5963 [0.5777, 0.6154] \\
Exact, exactly 3 & 0.6401 [0.6202, 0.6598] & 0.5746 [0.5534, 0.5958] & 0.5964 [0.5771, 0.6152] \\
Exact, up to 3 & 0.6448 [0.6248, 0.6644] & 0.5820 [0.5604, 0.6030] & 0.6096 [0.5896, 0.6292] \\
InternVL3.5-38B & 0.3634 [0.3438, 0.3832] & 0.3003 [0.2817, 0.3194] & 0.4114 [0.3951, 0.4282] \\
InternVL3.5-8B & 0.2910 [0.2724, 0.3103] & 0.2416 [0.2240, 0.2597] & 0.3618 [0.3450, 0.3785] \\
Qwen3-VL-8B & 0.3770 [0.3568, 0.3964] & 0.3111 [0.2917, 0.3298] & 0.4309 [0.4148, 0.4469] \\
Molmo2-8B & 0.3286 [0.3085, 0.3489] & 0.2747 [0.2557, 0.2941] & 0.3909 [0.3743, 0.4075] \\
LLaVA-OneVision-7B & 0.3312 [0.3134, 0.3492] & 0.2579 [0.2416, 0.2743] & 0.3971 [0.3825, 0.4119] \\
MiniCPM-V-4.5 & 0.3693 [0.3490, 0.3897] & 0.3081 [0.2888, 0.3276] & 0.4107 [0.3937, 0.4279] \\
\bottomrule\end{tabular}
\caption{Original shared-candidate controls, ambiguity-aware consensus. Means and 95\% task-bootstrap intervals; all rows use 1,327 tasks.}
\label{tab:matched12_0_0}
\end{table}
\begin{table}[htbp]\centering\scriptsize
\begin{tabular}{lrrr}\toprule
Method & Overcoverage & Hit@0.5 & Invalid \\\midrule
Template@3 & 0.5401 [0.5195, 0.5605] & 0.3828 [0.3572, 0.4092] & 0.0000 [0.0000, 0.0000] \\
Qwen3-VL-32B & 0.5231 [0.5015, 0.5446] & 0.3760 [0.3497, 0.4024] & 0.0008 [0.0000, 0.0023] \\
Greedy Oracle@3 & 0.1226 [0.1087, 0.1371] & 0.6330 [0.6074, 0.6594] & 0.0000 [0.0000, 0.0000] \\
Exact, exactly 3 & 0.1214 [0.1074, 0.1361] & 0.6330 [0.6066, 0.6586] & 0.0000 [0.0000, 0.0000] \\
Exact, up to 3 & 0.0636 [0.0535, 0.0742] & 0.6368 [0.6104, 0.6631] & 0.0000 [0.0000, 0.0000] \\
InternVL3.5-38B & 0.5408 [0.5191, 0.5627] & 0.3384 [0.3127, 0.3640] & 0.0211 [0.0136, 0.0294] \\
InternVL3.5-8B & 0.6124 [0.5902, 0.6345] & 0.2668 [0.2434, 0.2901] & 0.0580 [0.0460, 0.0708] \\
Qwen3-VL-8B & 0.5622 [0.5406, 0.5838] & 0.3519 [0.3255, 0.3775] & 0.0008 [0.0000, 0.0023] \\
Molmo2-8B & 0.6071 [0.5853, 0.6291] & 0.3044 [0.2796, 0.3293] & 0.0000 [0.0000, 0.0000] \\
LLaVA-OneVision-7B & 0.6608 [0.6419, 0.6791] & 0.3120 [0.2871, 0.3369] & 0.0000 [0.0000, 0.0000] \\
MiniCPM-V-4.5 & 0.5435 [0.5213, 0.5659] & 0.3429 [0.3173, 0.3693] & 0.0000 [0.0000, 0.0000] \\
\bottomrule\end{tabular}
\caption{Original shared-candidate controls, ambiguity-aware consensus. Means and 95\% task-bootstrap intervals; all rows use 1,327 tasks.}
\label{tab:matched12_0_1}
\end{table}
\begin{table}[htbp]\centering\scriptsize
\begin{tabular}{lrrr}\toprule
Method & Dice & IoU & B-F1 \\\midrule
Template@3 & 0.3989 [0.3801, 0.4176] & 0.3201 [0.3023, 0.3380] & 0.4348 [0.4190, 0.4506] \\
Qwen3-VL-32B & 0.3988 [0.3792, 0.4185] & 0.3291 [0.3103, 0.3481] & 0.4383 [0.4219, 0.4549] \\
Greedy Oracle@3 & 0.6431 [0.6232, 0.6630] & 0.5816 [0.5605, 0.6028] & 0.6080 [0.5886, 0.6270] \\
Exact, exactly 3 & 0.6432 [0.6228, 0.6628] & 0.5818 [0.5602, 0.6026] & 0.6085 [0.5890, 0.6275] \\
Exact, up to 3 & 0.6498 [0.6295, 0.6703] & 0.5913 [0.5696, 0.6132] & 0.6253 [0.6052, 0.6456] \\
InternVL3.5-38B & 0.3674 [0.3476, 0.3872] & 0.3059 [0.2871, 0.3249] & 0.4085 [0.3915, 0.4256] \\
InternVL3.5-8B & 0.2938 [0.2740, 0.3138] & 0.2465 [0.2281, 0.2657] & 0.3560 [0.3386, 0.3738] \\
Qwen3-VL-8B & 0.3798 [0.3597, 0.4008] & 0.3155 [0.2964, 0.3353] & 0.4279 [0.4113, 0.4451] \\
Molmo2-8B & 0.3296 [0.3092, 0.3504] & 0.2781 [0.2586, 0.2978] & 0.3852 [0.3681, 0.4026] \\
LLaVA-OneVision-7B & 0.3334 [0.3155, 0.3519] & 0.2615 [0.2452, 0.2782] & 0.3911 [0.3761, 0.4065] \\
MiniCPM-V-4.5 & 0.3686 [0.3485, 0.3888] & 0.3095 [0.2901, 0.3288] & 0.4063 [0.3891, 0.4237] \\
\bottomrule\end{tabular}
\caption{Original shared-candidate controls, strict consensus. Means and 95\% task-bootstrap intervals; all rows use 1,327 tasks.}
\label{tab:matched12_1_0}
\end{table}
\begin{table}[htbp]\centering\scriptsize
\begin{tabular}{lrrr}\toprule
Method & Overcoverage & Hit@0.5 & Invalid \\\midrule
Template@3 & 0.5613 [0.5410, 0.5813] & 0.3791 [0.3527, 0.4047] & 0.0000 [0.0000, 0.0000] \\
Qwen3-VL-32B & 0.5380 [0.5166, 0.5592] & 0.3828 [0.3564, 0.4099] & 0.0008 [0.0000, 0.0023] \\
Greedy Oracle@3 & 0.1492 [0.1334, 0.1650] & 0.6383 [0.6127, 0.6639] & 0.0000 [0.0000, 0.0000] \\
Exact, exactly 3 & 0.1486 [0.1325, 0.1646] & 0.6383 [0.6119, 0.6639] & 0.0000 [0.0000, 0.0000] \\
Exact, up to 3 & 0.0779 [0.0663, 0.0898] & 0.6436 [0.6172, 0.6692] & 0.0000 [0.0000, 0.0000] \\
InternVL3.5-38B & 0.5522 [0.5305, 0.5736] & 0.3482 [0.3225, 0.3730] & 0.0211 [0.0136, 0.0294] \\
InternVL3.5-8B & 0.6231 [0.6010, 0.6452] & 0.2735 [0.2502, 0.2984] & 0.0580 [0.0460, 0.0708] \\
Qwen3-VL-8B & 0.5812 [0.5592, 0.6030] & 0.3602 [0.3346, 0.3866] & 0.0008 [0.0000, 0.0023] \\
Molmo2-8B & 0.6197 [0.5975, 0.6416] & 0.3067 [0.2826, 0.3316] & 0.0000 [0.0000, 0.0000] \\
LLaVA-OneVision-7B & 0.6713 [0.6527, 0.6896] & 0.3157 [0.2909, 0.3406] & 0.0000 [0.0000, 0.0000] \\
MiniCPM-V-4.5 & 0.5696 [0.5480, 0.5914] & 0.3474 [0.3225, 0.3723] & 0.0000 [0.0000, 0.0000] \\
\bottomrule\end{tabular}
\caption{Original shared-candidate controls, strict consensus. Means and 95\% task-bootstrap intervals; all rows use 1,327 tasks.}
\label{tab:matched12_1_1}
\end{table}
\paragraph{Paired comparisons.}
On the ambiguity-aware target, Qwen3-VL-32B minus Template@3 (12) has mean
Dice $-.00697$ with 95\% interval $[-.02157,.00753]$; under strict consensus,
the difference is $-.00006$ with interval $[-.01450,.01435]$. Neither supports
a Dice advantage over the matched template. Full-bank minus 12-bank greedy
Dice is .03173 $[.02678,.03703]$; the template difference is .01317
$[.00824,.01828]$. Exactly-three exact minus greedy Dice is .00008
$[.00004,.00014]$, and allowing up to three improves the exact reference by
.00468 $[.00361,.00596]$. These isolate budget and selection effects rather
than changing the learned predictor. Complete paired intervals for all five
metrics are included with the result files.
Class results (\Cref{tab:matched12_class_0,tab:matched12_class_1}) show that no
deployable selector leads throughout, while oracle coverage varies
substantially with composition.

\begin{longtable}{lrrrrrrrrr}
\caption{Per-class Dice on the original 12 candidates, ambiguity-aware consensus. H: horizontal; V: vertical; D: diagonal; C: curved; T: triangle; R: rule of thirds; Ce: center; S: symmetric; P: pattern.}\label{tab:matched12_class_0}\\
\toprule Method & H & V & D & C & T & R & Ce & S & P \\\midrule\endfirsthead
\toprule Method & H & V & D & C & T & R & Ce & S & P \\\midrule\endhead
Template@3 & 0.566 & 0.340 & 0.354 & 0.291 & 0.408 & 0.334 & 0.457 & 0.307 & 0.213 \\
Qwen3-VL-32B & 0.571 & 0.544 & 0.356 & 0.281 & 0.301 & 0.363 & 0.437 & 0.302 & 0.184 \\
Greedy Oracle@3 & 0.839 & 0.716 & 0.581 & 0.509 & 0.563 & 0.858 & 0.655 & 0.447 & 0.266 \\
Exact, exactly 3 & 0.840 & 0.716 & 0.581 & 0.509 & 0.563 & 0.858 & 0.655 & 0.447 & 0.266 \\
Exact, up to 3 & 0.842 & 0.717 & 0.585 & 0.514 & 0.567 & 0.865 & 0.664 & 0.447 & 0.267 \\
InternVL3.5-38B & 0.552 & 0.376 & 0.339 & 0.302 & 0.282 & 0.236 & 0.429 & 0.281 & 0.161 \\
InternVL3.5-8B & 0.517 & 0.245 & 0.294 & 0.248 & 0.199 & 0.186 & 0.311 & 0.188 & 0.158 \\
Qwen3-VL-8B & 0.590 & 0.436 & 0.367 & 0.300 & 0.312 & 0.319 & 0.378 & 0.281 & 0.181 \\
Molmo2-8B & 0.556 & 0.285 & 0.276 & 0.262 & 0.233 & 0.270 & 0.391 & 0.186 & 0.156 \\
LLaVA-OneVision-7B & 0.548 & 0.295 & 0.331 & 0.251 & 0.262 & 0.338 & 0.313 & 0.247 & 0.159 \\
MiniCPM-V-4.5 & 0.543 & 0.398 & 0.351 & 0.278 & 0.270 & 0.398 & 0.395 & 0.279 & 0.173 \\
\bottomrule\end{longtable}
\clearpage
\begin{longtable}{lrrrrrrrrr}
\caption{Per-class Dice on the original 12 candidates, strict consensus. H: horizontal; V: vertical; D: diagonal; C: curved; T: triangle; R: rule of thirds; Ce: center; S: symmetric; P: pattern.}\label{tab:matched12_class_1}\\
\toprule Method & H & V & D & C & T & R & Ce & S & P \\\midrule\endfirsthead
\toprule Method & H & V & D & C & T & R & Ce & S & P \\\midrule\endhead
Template@3 & 0.566 & 0.357 & 0.327 & 0.277 & 0.406 & 0.339 & 0.460 & 0.309 & 0.217 \\
Qwen3-VL-32B & 0.568 & 0.557 & 0.366 & 0.279 & 0.302 & 0.368 & 0.443 & 0.298 & 0.207 \\
Greedy Oracle@3 & 0.840 & 0.718 & 0.588 & 0.474 & 0.569 & 0.865 & 0.660 & 0.448 & 0.296 \\
Exact, exactly 3 & 0.840 & 0.718 & 0.588 & 0.474 & 0.570 & 0.866 & 0.661 & 0.448 & 0.296 \\
Exact, up to 3 & 0.843 & 0.721 & 0.599 & 0.490 & 0.576 & 0.873 & 0.669 & 0.448 & 0.296 \\
InternVL3.5-38B & 0.553 & 0.382 & 0.348 & 0.291 & 0.285 & 0.241 & 0.436 & 0.278 & 0.177 \\
InternVL3.5-8B & 0.525 & 0.248 & 0.303 & 0.210 & 0.203 & 0.191 & 0.315 & 0.192 & 0.166 \\
Qwen3-VL-8B & 0.584 & 0.438 & 0.383 & 0.283 & 0.320 & 0.324 & 0.381 & 0.278 & 0.193 \\
Molmo2-8B & 0.559 & 0.289 & 0.286 & 0.218 & 0.236 & 0.275 & 0.395 & 0.188 & 0.163 \\
LLaVA-OneVision-7B & 0.549 & 0.302 & 0.336 & 0.226 & 0.266 & 0.343 & 0.316 & 0.244 & 0.183 \\
MiniCPM-V-4.5 & 0.534 & 0.421 & 0.342 & 0.248 & 0.272 & 0.403 & 0.401 & 0.276 & 0.195 \\
\bottomrule\end{longtable}